%% file: paper.tex
\documentclass[sigconf]{acmart}

\setcopyright{none}
\renewcommand\footnotetextcopyrightpermission[1]{}

\usepackage{amsmath}
\usepackage{amsfonts}
\usepackage{algorithmicx}
\usepackage{bbding}
\usepackage{nopageno}
\usepackage{amsthm}
\usepackage{algorithm}
\usepackage[noend]{algpseudocode}
\usepackage{graphicx}
\usepackage{textcomp}
\usepackage{multirow}
\usepackage{enumitem}
\usepackage{pifont}
\usepackage{xcolor}
\usepackage{bm}
\usepackage{array}
\usepackage{ragged2e}
\newcolumntype{P}[1]{>{\centering\arraybackslash}p{#1}}
\newcolumntype{M}[1]{>{\centering\arraybackslash}m{#1}}

\newcommand\checkgpt{\textsf{CheckGPT}}

\definecolor{Blue}{RGB}{0,0,255}
\newcommand{\newc}{\color{blue}}

\newcommand{\hic}{\color{red}}

\DeclareMathOperator*{\argmin}{\arg\!\min}

\begin{document}

\title[On the Detectability of ChatGPT Content: Benchmarking, Methodology, and Evaluation through the Lens of Academic Writing]{On the Detectability of ChatGPT Content: \\Benchmarking, Methodology, and Evaluation through \\ the Lens of Academic Writing}


\author{Zeyan Liu, Zijun Yao, Fengjun Li, and Bo Luo}
\affiliation{%
  \institution{EECS/I2S, The University of Kansas}
  \city{Lawrence}
  \state{KS}
  \country{USA}}
\email{{zyliu, zyao, fli, bluo}@ku.edu}

\begin{abstract}
With ChatGPT under the spotlight, utilizing large language models (LLMs) to assist academic writing has drawn a significant amount of debate in the community. In this paper, we aim to present a comprehensive study of the detectability of ChatGPT-generated content within the academic literature, particularly focusing on the abstracts of scientific papers, to offer holistic support for the future development of LLM applications and policies in academia. Specifically, we first present GPABench2, a benchmarking dataset of over 2.8 million comparative samples of human-written, GPT-written, GPT-completed, and GPT-polished abstracts of scientific writing in computer science, physics, and humanities and social sciences.  Second, we explore the methodology for detecting ChatGPT content. We start by examining the unsatisfactory performance of existing ChatGPT detecting tools and the challenges faced by human evaluators (including more than 240 researchers or students). We then test the hand-crafted linguistic features models as a baseline and develop a deep neural framework named \checkgpt~ to better capture the subtle and deep semantic and linguistic patterns in ChatGPT written literature.
Last, we conduct comprehensive experiments to validate the proposed \checkgpt~ framework in each benchmarking task over different disciplines. To evaluate the detectability of ChatGPT content, we conduct extensive experiments on the transferability, prompt engineering, and robustness of \checkgpt.

\end{abstract}




\maketitle

\input{tex/intro}

\input{tex/datacollection}

\input{tex/survey}

\input{tex/checkgpt}
\input{tex/evaluation}

\input{tex/background}
\input{tex/conclusion}



\bibliographystyle{ACM-Reference-Format}
\bibliography{tex/ref}

\input{tex/appendix}

\end{document}

%% file: tex/intro.tex
\section{Introduction}
\label{sec:intro}

The recently debuted Large Language Model (LLM) - ChatGPT has shown an impressive ability to generate sophisticated texts with human-like language style and quality. Concerns have been raised that the LLM-generated content (LLM-content) can be misused to abuse the trust systems we have, e.g., in cheating and plagiarism \cite{stokel2022ai,khalil2023will}, or in phishing and romance scams \cite{roy2023generating,grbic2023social}. While many academic institutes and publishers have announced policies on the usage of LLM content, it is hard to enforce such policies unless we have a tool to accurately detect LLM content.

LLM-content detection can be challenging due to the unique characters of LLM/ChatGPT: (1) like a human conversationalist, the output of LLM has a relevant, organized response with a low level of grammar errors; (2) the sampling mechanism of LLM output ensures that the choice of words is stochastic, therefore, the responses are distinct even with multiple repeated inquiries; and (3) the misuse of LLM-content can be stealthy since users can invoke ChatGPT to polish human writing. Facing these challenges, existing LLM detectors perform deficiently, especially in detecting GPT-polished text (Sec. \ref{sec:comparison}). Some experiences in identifying LLM-content have been reported in the literature, e.g., \citet{guo2023close} and \citet{liu2023argugpt} noted that ChatGPT output tends to be more objective, formal, focused, and fluent than human-content. However, a holistic investigation of the distinguishability of LLM-Content is still missing. 

To this end, in this paper, we first identify three typical cases of using or abusing ChatGPT in academic writing: {\em composing}, {\em completing}, and {\em polishing}. We pick three representative disciplines for investigation: \textit{computer science} for technical/engineering writing, \textit{physics} for science writing, and \textit{humanities and social sciences} for liberal arts writing. To address a range of complex real-world scenarios, we used four different prompt patterns for each task across each discipline and collected a dataset, \textit{GPABench2}, with 2.8 million human-written and ChatGPT-generated academic abstracts. 

Next, we conducted an extensive field study with human evaluators to assess if they can distinguish LLM content accurately provided with a mixture of true and false samples. The cohort of 155 evaluators consisting of university faculty, researchers, and graduate students, proves that the recognition of LLM-content is difficult for visual inspection based on language appearance, with or without individual experiences of writing research articles. A second cohort of 87 evaluators were provided with ground truth data as references: pairs of human-written and GPT-generated abstracts. While their capacities to identify human- and GPT-generated text were improved, the accuracy is still too low for reliable detection. In addition, we test 10+ state-of-the-art online or open-source detectors on GPABench2, e.g., GPTZero, and show that they demonstrate modest to poor performance, except for the language-model-based detectors, like BERT and RoBERTa, which archives significantly better result but requires excessive training efforts. 

Last, we develop and evaluate a language-model-based detection framework, named \checkgpt, to explore the feasibility of building automated tools for LLM-content detection in a niche area. \checkgpt~has the following advantages: (1) it is a black-box solution that leverages deep learning frameworks to achieve a high accuracy compared to human and state-of-the-art (SOTA) LLM-content detectors. 
(2) \checkgpt~adopts a model-agnostic setting that it can be treated as a plugin to most pre-trained language models (e.g., BERT), as a result, the number of parameters to be trained can be largely reduced. (3) Due to its ability to learn generalized semantic patterns of LLM-content, \checkgpt~shows a strong potential for domain transfer that only requires minimum fine-tuning efforts. 
Finally, we conduct comprehensive experiments to demonstrate \checkgpt's design goals and strengths: its performance on the GPABench2 dataset, its transferability to new domains, new models, and new prompts, and its robustness against continuous updates of ChatGPT and against human-modified GPT writing.

In summary, our main contributions are:

\begin{itemize}[itemsep=0.5pt,topsep=0.5pt,leftmargin=*]
\item We present 
GPABench2, a cross-disciplinary corpus consisting of human-written, GPT-written, GPT-completed, and GPT-polished research paper abstracts. GPABench2 has the potential to serve as a cornerstone for benchmarking GPT detectors, and a valuable resource to assist in the design of new detecting methods.
\item We evaluate the SOTA GPT detectors with GPABench2 and present their performance. {Meanwhile, with a user study of 242 participants, we show that human evaluators are unable to distinguish between human-written and GPT-generated academic writing. This incapability persists regardless of experience, knowledge, and reference. }
\item We present \checkgpt, a deep-learning-based and model-agnostic GPT-content detector with validated benefits of affordability, transferability, and Robustness\footnote{We share \checkgpt~at \url{https://anonymous.4open.science/r/CheckGPT-80B2}}.
We demonstrate the outstanding performance ($\sim$99\% average accuracy) of \checkgpt~with extensive experiments.
\end{itemize}

\vspace{1mm}\noindent\textbf{{Ethical Considerations.}} The user study in Section \ref{sec:survey} was reviewed and approved by the Human Research Protection Program at the University of [Anonymized]. 
All the research paper abstracts collected in Section \ref{sec:gpabenchmark} are open to the public. We invoked ChatGPT's API (with payment) to collect the GPT-generated abstracts. The GPABench2 dataset and the \checkgpt~tool have been open-sourced. The academic community is actively discussing how AI writing assistance tools may pose challenges to research and education \cite{sallam2023chatgpt,malinka2023educational,anders2023using}. OpenAI also posted their perspectives on the education-related risks and opportunities \cite{openai2023educator}. 


The rest of the paper is organized as follows: 
We introduce the GPABench2 dataset and evaluate the open-source and commercial ChatGPT detectors in Section \ref{sec:gpabenchmark}, followed by a user study of ChatGPT-content detection in Section \ref{sec:survey}. We present the technical details of \checkgpt~in Section \ref{sec:checkgpt} and experimental results in Section \ref{sec:exp}. 
Finally, we survey the related literature in Section \ref{sec:background}, and conclude the paper in Section \ref{sec:conclusion}.

%% file: tex/datacollection.tex
\section{GPABenchmark: \underline{G}PT Cor\underline{p}us for \underline{A}cademia}\label{sec:gpabenchmark}

\subsection{The GPABench2 Dataset}\label{sec:subsecgpabenchmark}

The concern of LLM/ChatGPT misuse has been raised widely in academia because (1) academic integrity violations such as cheating and plagiarism will become {\em easy-to-conduct} and {\em hard-to-detect}. (2) False and redundant information may flood the publication systems. Stack Overflow had to ban LLM-generated posts to ensure that visitors can find reliable answers efficiently \cite{stackoverflowbanchatgpt} Academic conferences started to ban LLM-generated texts (e.g., ICML) or require disclosure (e.g., ACL, Nature, and  RSC). However, without effective mechanisms to detect GPT-generated content and benchmarking datasets to train/evaluate the detectors, such discussions and policies become meaningless. 

The state-of-the-art corpora for ChatGPT text classification mainly focus on question-and-answer (Q\&A) dialogues \cite{he2023mgtbench, guo2023close}. 
While the Q\&A datasets align with the original design of ChatGPT as an interactive ``Chat'' interface, they become insufficient as the usage scenarios of ChatGPT have significantly expanded beyond chat.   
When ChatGPT is adopted for academic writing, such as essays, reports, and even research papers \cite{firat2023chatgpt,stokel2022ai}, the generated text akin to academic writing style is often objective, formal, and focused \cite{guo2023close, liu2023argugpt}, posing more challenges to the detectors: (1) Human conversations often contain subjective opinions, personal biases, and emotions. However, such clues are significantly less observed in academic writing, which is generally formal and objective \cite{hyland2005stance, biber1991variation, biber2010challenging, guo2023close}. (2) Grammatical errors and inconsistencies in human-generated texts may serve as meaningful indicators, however, they are less likely to occur in academic writing, which is expected to meet higher standards for fluency, clarity, and grammatical correctness \cite{hinkel2003teaching, lynch2013grammar}.  Also, academic writers typically adopt a comprehensive and organized style \cite{cargill2021writing} akin to the one generated by ChatGPT \cite{guo2023close}. (3) Academic abstracts typically explore domain-specific and highly-specialized topics \cite{hyland2005stance}, which lead to a significantly different term distribution from conversational dialogues. 

With the unique characteristics of academic writing, a new ChatGPT-generated corpus is necessary for benchmarking GPT detectors and assisting in the design of detectors. 
In this paper, we introduce \textit{GPABench2} (GPABenchmark version 2), a large-scale \underline{G}PT-generated text cor\underline{p}us for \underline{a}cademic writing. 

We first collected research papers (title and abstracts) from three disciplines: \textit{computer science} (CS) abstracts from top-tier conference proceedings and arXiv, \textit{physics} (PHX) from arXiv, and \textit{humanities and social sciences} (HSS) from Springer's SSRN including history, philosophy, sociology, and psychology disciplines. The three fields spread across ``hard science'' (math-intensive) and ``soft science'' disciplines. For CS and Physics, we chose papers published or posted on or before 2019 (before the release of GPT-3) to ensure
that they were all human-written, as researchers may have adopted GPT-3 to assist their writings before the web-based ChatGPT was released\footnote{GPT-3 was first released in 06/2020, access to the test release was by-invitation-only until 11/2021 when the API was made publicly accessible.}. 
Eventually, we collected 50,000 papers from each discipline. 


We define three tasks based on the most representative scenarios where LLMs are used/misused in academic writing: 

\begin{itemize}[noitemsep, topsep=1mm, left=0pt]
\item\textbf{Task 1. GPT-written full abstracts (GPT-WRI or WRI).} The author gives a title to ChatGPT and asks it to write the complete abstract from scratch. 


\item\textbf{Task 2. GPT-completed abstracts (GPT-CPL or CPL).} Text completion is considered a conventional function of LLMs: The author provides a few sentences to ChatGPT, who follows the logic to complete the rest of the paragraph. We mimic this scenario: for an abstract with $s$ sentences, the first $s/2$ sentences are provided to ChatGPT, based on which it \textit{completes the abstract with $w$ words}, where $w$ is the word count in the second half of the original abstract. Hence, the GPT-CPL abstract will have \textit{approximately} the same length as the human-written abstract. 

\item\textbf{Task 3. GPT-polished abstracts (GPT-POL or POL).} We provide the entire abstract to ChatGPT for polishing. 
ChatGPT re-writes the text sentence-by-sentence and generates a revised abstract. Invoking ChatGPT multiple times will generate different results for the same seed abstract. 
\end{itemize}


In this paper, we use \textit{human-written} abstracts (HUM) to denote abstracts that are completely written by human authors. We use \textit{GPT-Generated} abstract (GPT-GEN) to denote all three categories of GPT-content described above: GPT-WRI, -CPL, and -POL.

We applied prompt engineering in data collection to ensure a broad coverage of ChatGPT use cases. 
We studied popular prompt patterns \cite{white2023prompt} and prompt guidelines \cite{PlexPt2023prompt, Akın2023prompt, amiri2023prompt, jaiswal2023prompts} in the literature and crafted four distinct prompts for each task, denoted as Prompts 1 to 4 (presented in Appendix~\ref{apdx:prompt3}): Prompt 1 is a popular but straightforward zero-shot prompt. Prompt 2 integrates the contexts to outline the scope of a specific discipline. Prompt 3 uses the role-playing technique to specify a ``persona'', e.g., ``an expert paper writer in computer science''. Prompt 4 provides detailed requirements and instructions to guide ChatGPT. These prompts represent four use cases with increasing levels of knowledge fed to ChatGPT. 

We invoke ChatGPT (gpt-3.5-turbo) through OpenAI's API to generate abstracts at the cost of 0.2 cents per 1,000 tokens. In three months, we collected 50,000 samples for each prompt, task, and discipline, as the GPABench2 Main Dataset (1.8 million total GPT-GEN samples). We further adopted ten advanced prompting techniques, e.g., chain-of-thought and in-context prompt learning, to generate 435K additional testing samples (Sec. \ref{sec:discussion}). Eventually, GPABench2 contains 2.385M total samples (2.235M GPT-GEN and 0.15M HUM).

\vspace{2mm}\noindent\textbf{Advanced Prompt Engineering and Additional Testing Data.} Research efforts on prompt engineering aim to guide or improve the design of ChatGPT prompts \cite{white2023prompt,ekin2023prompt}. We adopt six approaches that are widely adopted in the community: (1) Zero-shot Chain-of-Thought Prompting (ZC, \cite{zeroshot}) enforces step-by-step reasoning with specific trigger phrases like ``Let's think step by step.'' (2) Automatic Prompt Engineer (APE, \cite{ape}) automates the creation and selection of prompts using iterative optimization. (3) Self-critique Prompting (SCP, \cite{critique}) employs GPT to evaluate its own responses and provide feedback. (4) Few-shot Prompting (FSP, \cite{fewshot}) conditions the model using examples or demonstrations. (5) Least-to-Most Prompting (LMP, \cite{least}) parses a problem into simpler subproblems. (6) Generated Knowledge Prompting (GKP, \cite{generated}) starts the prompt with relevant information generation. We also adopt four prompt refinement methods: (1) Prompt Perfect (PP, \cite{promptperfect}). (2) GPT-generated Prompts (GP, \cite{Creator}). (3) Meta Prompts (MP, \cite{metaprompt}). (4) Instruction Induction (II, \cite{honovich2022instruction}, not for Task 3). We use each method to write, complete, and polish 5,000 abstracts from each discipline (please refer to Appendix~\ref{apdx:otherprompt} for details). In summary, we collected 435K additional samples to be used for testing.

\begin{table}[t]
	\setlength{\tabcolsep}{0.38em}
 \caption{Performance of commercial GPT detectors on GPABench2. Red: detection accuracy <50\%, or average score on the wrong side of the decision threshold. T1/2/3: Task 1/2/3. GPT-WRI/CPL/POL: GPT-written/completed/polished abstracts.}\label{tab:otherdetectors} \vspace{-2mm}
    {
	\begin{tabular}{c|ccc|ccc|ccc}
        \hline
        
      \multirow{2}{*}{} & \multicolumn{3}{c|}{T1. GPT-WRI} &\multicolumn{3}{c|}{T2. GPT-CPL}  &\multicolumn{3}{c}{T3. GPT-POL}\\
		\cline{2-10}
        & CS & PHX & HSS & CS & PHX & HSS & CS & PHX & HSS\\\hline\hline
        \multicolumn{10}{c}{(a) Classification accuracy (in \%) of GPTZero.} \\\hline
GPT & \hic{30.3}	&	\hic{25.3}	&	72.0	&	\hic{17.0}	&	\hic{6.0}	&	\hic{43.7}	&	\hic{1.7}	&	\hic{2.3}	& \hic{20.3}	\\
Human	& 99.3	&	99.7 &	100   &	99.7	&	99.7	&	94.3	&	99.7	&	95.7	&	95.7	\\\hline\hline
    \multicolumn{10}{c}{(b1) Detection accuracy (in \%) of ZeroGPT} \\\hline
GPT & 67.4	&	68.4	&	92.3	&	\hic{25.3}	&	\hic{10}	&	62.4	&	\hic{3.3}	&	\hic{2.7}	&	\hic{24.7}	\\
Human	& 100	&	98.4	&	95	&	99.7	&	99.7	&	94.7	&	98.3	&	98.6	&	92.7 \\ \hline
    \multicolumn{10}{c}{(b2) Average score reported by ZeroGPT. 0:human, 8:GPT} \\\hline
GPT & 5.43	&	5.39	&	7.41	&	\hic{2.26}	&	\hic{0.97}	&	4.97	&	\hic{0.35}	&	\hic{0.29}	& \hic{2.15}	\\
Human	& 0.09	&  0.13 &	0.52   &	0.08	&	0.04	&	0.47	&	0.20	&	0.14	&	0.64	\\ \hline\hline
    \multicolumn{10}{c}{(c1) Detection accuracy (in \%) of OpenAI's detector} \\\hline
GPT & 80.7	 & 	70	 & 	63	 & 	63.7	 & 	\hic{23.7}	 & 	\hic{27.3}	 & 	\hic{6.3}	 & 	\hic{4.3}	 & 	\hic{6}  \\
Human	& 51.0	 & 	69.7	 & 	84.0	 & 	\hic{35.3}	 & 	59.7	 & 	79.6	 & 	50.7	 & 	69.0	 & 	88.0 \\
    \hline
    \multicolumn{10}{c}{(c2) Average score reported by OpenAI. 0:human, 4:GPT} \\\hline
GPT & 3.11	&	2.89	&	2.72	&	2.70	&	2.12	&	2.04	&	\hic{1.75}	&	\hic{1.59}	& \hic{1.52}	\\
Human	& 1.42	&  1.17 &	0.59   &	1.71	&	1.35	&	0.68	&	1.38	&	1.14	&	0.52	\\
    \hline   
        \end{tabular}
        }\vspace{-3mm}
\end{table}

\subsection{Benchmarking Online and Open-source ChatGPT Detectors}\label{sec:benchmark}\label{sec:comparison}

Several online commercial tools have been developed to detect AI-generated text. We are especially interested in them because they are the most accessible and easy-to-use detectors for ordinary users. We evaluate the accuracy of three representative online ChatGPT tools, GPTZero \cite{gptzero}, ZeroGPT \cite{zerogpt}, and OpenAI's classifier \cite{openaidetector}, over GPABench2. Due to a lack of API, slow responses, and high cost, we cannot run large-scale experiments. Instead, we randomly sampled 300 pairs of human-written and the corresponding GPT-generated abstracts for each task in each discipline, i.e., 2,400 pairs in total, and fed them to each detector. Their performance is summarized in Table \ref{tab:otherdetectors}. Note that, in Task 2 (GPT-completed abstracts), we only submitted the second half of each abstract to the detectors.  

From the performance summary in Table \ref{tab:otherdetectors} and the detailed results in Appendix \ref{apdx:benchmarking}, we have three observations: (1) all three detectors demonstrated modest to poor accuracy for GPT-GEN content; (2) the detectors have tendencies to classify GPT-generated text as human-written; and (3) the detection accuracy for GPT-GEN decreases significantly from Task 1 (GPT-WRI) to Task 3 (GPT-POL). Note that this experiment is not intended for comparison with CheckGPT. Since these models are not explicitly trained with academic data, their inaccuracy can be excused and directly comparing them with CheckGPT is unfair. Rather, we intend to use the results as a motivation for CheckGPT--generic detectors struggle in specific tasks, indicating limited transferability. The gap highlights the need for effective detectors for this niche domain with the potential to transfer to related domains. 

A number of detectors have been proposed for LLM/ChatGPT-generated text. For a comparative study, we adopted 15 open-source detectors in the literature. 
Based on the design philosophy, we further categorize them into three groups: (a) \textit{Pre-trained detectors}. We directly adopt the trained models: HC3-Perplexity (HC3-PPL) \cite{guo2023close}, HC3-GLTR \cite{guo2023close}, HC3-Roberta (HC3-RBT) \cite{guo2023close}, OpenAI-Roberta (OpenAI-RBT) \cite{solaiman2019release, openairoberta}. (b) \textit{Statistics-based detectors}. They analyze the statistical differences between LLM and human-written text: Histogram-of-Likelihood Ranks (HLR) \cite{gehrmann2019gltr}, Rank \cite{mitchell2023detectgpt}, Log-Rank \cite{mitchell2023detectgpt}, Total Probability (TP) \cite{solaiman2019release} Perplexity (PPL) \cite{guo2023close}, Entropy \cite{gehrmann2019gltr, mitchell2023detectgpt}, DetectGPT \cite{mitchell2023detectgpt}.  (c) \textit{Fine-tuned Language Models}. They fine-tune the pre-trained language models for detection: BERT \cite{ippolito2020automatic}, DistillBERT \cite{he2023mgtbench}, and RoBERTa \cite{liu2023argugpt, guo2023close, wang2023m4, macko2023multitude}. We also include GPT-2, which is even larger with 355M parameters. Please refer to Appendix \ref{appdx:detectordetails} for more details. 

The statistics- and fine-tuning-based detectors are trained/tuned with the entire training set in GPABench2. We evaluate all the detectors on GPABench2 with samples generated by Prompt 1. From the results presented in Table~\ref{tab:opensourcedetector}, we observe the following: (1) The pre-trained detectors work poorly on GPABench2. The failure of the OpenAI detector can be attributed to the discrepancy in the target model, as it was designed for GPT-2. For HC3 detectors, their poor performance indicates the ineffectiveness of statistics-based methods (HC3-PPL, HC3-GLTR) and the limited transferability of RoBERTa (HC3-RBT). (2) The statistics-based detectors provide satisfactory performance in Task 1, however, they become mostly ineffective in Tasks 2 and 3. These results also indicate that GPT-polished content demonstrates significantly more statistical similarities to human-written text, especially at the lexicon level, e.g., word distributions, complexity, etc. (3) Fine-tuned language models with a native classification layer provide outstanding performance in most tasks, with slightly lower accuracy in Task 3. However, training or tuning the full BERT, RoBERTa, or GPT models is computationally expensive, e.g., it takes over 1,000 seconds to fine-tune BERT or RoBERTa for one epoch on an NVidia 4090 GPU, and more than 6,000 seconds for GPT-2. 


\begin{table}[t]
    \caption{Performance (F1-score) of open-source detectors on GPABench2. Values in blue: the best performance.}\label{tab:opensourcedetector}
    \setlength{\tabcolsep}{0.27em}
    \centering
    \vspace{-1mm}
    {\small 
	\begin{tabular}{c|ccc|ccc|ccc}
 \hline
        & \multicolumn{3}{c|}{T1. GPT-WRI} &\multicolumn{3}{c|}{T2. GPT-CPL}  &\multicolumn{3}{c}{T3. GPT-POL}\\\hline
         & CS & PHX & HSS & CS & PHX & HSS & CS & PHX & HSS\\\hline
           \multicolumn{10}{c}{(a) Pre-trained Detectors} \\\hline
        HC3-PPL & 0.760 & 0.794 & 0.854 & 0.688 & 0.686 & 0.763 & 0.665 & 0.668 & 0.682 \\
        HC3-GLTR  & 0.679 & 0.680 & 0.735 & 0.670 & 0.669 & 0.700 & 0.666 & 0.667 & 0.670 \\
        HC3-RBT & 0.710 & 0.788 & 0.795 & 0.726 & 0.753 & 0.786 & 0.722 & 0.746 & 0.805 \\
        OpenAI-RBT & 0.072 & 0.106 & 0.110 & 0.159 & 0.219 & 0.192 & 0.072 & 0.113 & 0.112 \\ \hline
        \multicolumn{10}{c}{(b) Statistics-based Detectors} \\\hline
        HLR & 0.910 & 0.917 & 0.902 & 0.792 & 0.757 & 0.841 & 0.608 & 0.602 & 0.682 \\
        Rank & 0.781 & 0.650 & 0.525 & 0.783 & 0.612 & 0.531 & 0.808 & 0.738 & 0.620\\
        Log-Rank & 0.911 & 0.919 & 0.895 & 0.794 & 0.764 & 0.843 & 0.626 & 0.608 & 0.679 \\
        TP & 0.913 & 0.924 & 0.894 & 0.797 & 0.778 & 0.847 & 0.632 & 0.622 & 0.686 \\
        PPL & 0.913 & 0.925 & 0.891 & 0.814 & 0.797 & 0.848 & 0.654 & 0.650 & 0.695\\
        Entropy & 0.803 & 0.851 & 0.745 & 0.697 & 0.708 & 0.738 & 0.558 & 0.593 & 0.633 \\
        DetectGPT & 0.790 & 0.722 & 0.768 & 0.685 & 0.693 & 0.766 & 0.616 & 0.588 & 0.630 \\ \hline
        \multicolumn{10}{c}{(c) Fine-tuned Language Models} \\\hline
        BERT & 0.999 & 0.999 & 0.998 & 0.992 & 0.983 & 0.992 & 0.983 & 0.984 & 0.966 \\
        DistillBERT & 0.999 & 0.999 & {\newc 0.999} & 0.990 & 0.992 & 0.975 & 0.977 & 0.989 & 0.973 \\
        RoBERTa & {\newc 0.999} & 0.999 & 0.997 & 0.970 & 0.995 & {\newc 0.995} & 0.981 & 0.993 & 0.967 \\ 
        GPT2 & 0.998 & 0.995 & 0.998 & 0.982 & 0.987 & 0.947 & 0.969 & 0.974 & 0.956 \\ 
        \hline
        CheckGPT & 0.999 & {\newc 1.000} & {\newc 0.999} & {\newc 0.996} & {\newc 0.995} & {\newc 0.995} & {\newc 0.993} & {\newc 0.994} & {\newc 0.993} \\ \hline        
        \end{tabular}
        } \vspace{-4mm}
\end{table}

%% file: tex/survey.tex
\section{User Study: Identification of Human- and GPT-Generated Abstracts}\label{sec:survey}

With all the news reports and online/informal discussions that human users are unable to distinguish ChatGPT-generated text from man-written text, we investigate this problem through a user study in a relatively well-defined domain: research publications. We aim to answer four research questions: 

\vspace{3pt}
\noindent
\textbf{RQ1:} Could (experienced) researchers distinguish between human-written and GPT-generated paper abstracts?

\vspace{3pt}
\noindent
\textbf{RQ2:} Do prior experiences with reading/writing papers contribute to the capability of identifying GPT-generated abstracts? 

\vspace{3pt}
\noindent
\textbf{RQ3:} Does the researchers' capability in identifying GPT content 
vary by discipline? 

\vspace{3pt}
\noindent
\textbf{RQ4:} Will their capability improve if they have ground truth data (pairs of human- and GPT-generated content) as references?
\vspace{3pt}

We designed a questionnaire as follows: 
first, the landing page displays an IRB information statement and asks the participants to select their ``most familiar discipline'' among CS, Physics, and Humanities \& Social Sciences (HSS). Then, the main questionnaire page asks the participants to provide basic background information, their roles, whether they have published research papers, and self-claimed familiarity with research papers. Finally, each participant is presented with three abstracts and asked to annotate each as ``human-written'' or ``GPT-GEN/POL''. Each abstract is randomly sampled from HUM or GPT-WRI/POL abstracts from Tasks 1 and 3 of GPABench2. For Task 3, we display the following hint: ``This abstract was completely written by humans OR written by humans and then polished by ChatGPT.'' 

In the second experiment, we provide the participant with some ground truth data as reference, i.e., we display three pairs of labeled human- and GPT-generated abstracts from the same task and the same discipline, and instruct the participants to learn the writing styles from these abstracts before moving to the detection questions. Three unlabeled samples are then displayed to the user to annotate. 

\begin{table}[t]
	\caption{Results of the user study. Par.: number of participants; Acc.: accuracy; HUM: accuracy for human-written abstracts; GPT: accuracy for GPT-generated abstracts.}\label{tab:userstudy1}
 \vspace{-4mm}
    \setlength{\tabcolsep}{2pt}
    \centering
    {
	\begin{tabular}{c|cccc|cccc}
        \hline
        & \multicolumn{4}{c|}{Exp 1: w/o reference} & \multicolumn{4}{c}{Exp 2: with reference} \\ \hline
	      Category & Par. & Acc. & HUM & GPT  & Par. & Acc. & HUM & GPT \\\hline
        \multicolumn{9}{c}{Role}\\\hline
Faculty & 	44 & 49.2\% & 	58.6\% & 	41.9\%  & 28 & 59.5\% & 61.2\% & 57.1\% \\
Researchers & 30 & 50.0\% & 	58.2\% & 	37.1\% & 28 & 58.3\% & 	56.3\% & 	61.1\% \\
Students & 	81 & 48.1\% & 	56.3\% & 	40.3\%  & 31 & 66.7\% & 	75.6\% & 	58.3\% \\ 
        \hline
        \multicolumn{9}{c}{Discipline}\\\hline
CS & 	57 & 50.3\% & 	59.0\% & 	43.0\% & 33 & 60.6\% & 	59.6\% & 61.5\%\\
Physics & 	48 &	53.5\% & 	65.1\% & 37.7\% & 25 & 60.0\% & 	71.1\% & 	43.3\%\\
HSS & 	50 & 42.7\% & 	46.5\% & 	39.2\% & 29 & 64.4\% & 	62.0\% & 	67.6\% \\
        \hline
        \multicolumn{9}{c}{Self-claimed Familiarity with Research Papers}\\\hline
Expert & 	52  & 	51.3\% & 	60.6\% & 	43.5\%  & 32 & 61.5\% & 	67.3\% & 	53.7\% \\
Knowledgable & 	56 & 	47.6\% & 	57.3\% & 	34.7\% & 30 & 61.1\% & 	56.5\% & 	65.9\% \\
Somewhat & 	39 & 48.7\% & 	56.0\% & 	43.3\% & 21 & 65.1\% & 	72.7\% & 56.7\% \\
No familiarity & 8  & 	41.7\% & 	46.7\% & 	33.3\% & 4 & 50.0\% & 	50.0\% & 	50.0\% \\
        \hline
        \multicolumn{9}{c}{Published papers?}\\\hline
Yes & 	106 & 48.7\% & 	58.1\% & 	39.2\% & 56 & 60.7\% & 	64.0\% & 	57.0\% \\
No & 	49 & 49.0\% & 	55.6\% & 	42.7\% & 31 & 63.4\% & 	64.2\% & 	62.5\% \\
        \hline
        \end{tabular}} \vspace{-4mm}
\end{table}

We distributed questionnaires to faculty members, researchers, and graduate students in the Department of EECS, Department of Physics, and College of Liberal Arts at our University (in the US). Physics faculty members also shared the questionnaire with collaborators in a large research organization in Europe. For Experiment 1, we received 155 responses with 465 annotated abstracts in approximately four weeks. The overall accuracy, defined as the proportion of correctly identified abstracts out of all abstracts, was 48.82\%, which is slightly worse than random guesses. For Experiment 2, we received 87 responses with 261 annotated abstracts in two weeks. The overall accuracy improved to 61.69\%. Note that the median time spent on each questionnaire was 108 seconds in Experiment 1 and 229 seconds in Experiment 2, showing that the participants did spend time reading and comprehending the ground truth samples. The detailed statistics of the responses are shown in Table~\ref{tab:userstudy1}. From the responses, we have the following observations:

\noindent$\bullet~$ It is very challenging for human users to distinguish between human-written and GPT-generated paper abstracts. Only 21 users in Experiment 1 and 24 users in Experiment 2 correctly identified all three abstracts. If all participants were making random selections, 19.38 users would have scored 3 correct selections in Experiment 1. 

\noindent$\bullet~$ Participants have the tendency to annotate abstracts as ``human''. 57.33\% and 64.08\% of human-written abstracts were correctly labeled as ``human-written" in Experiments 1 and 2, respectively, while 59.66\% and 41.2\% of GPT-generated abstracts were mistakenly labeled as ``human-written''. The result confirms the public opinion that ChatGPT achieves human-like language style and quality. 

\noindent$\bullet~$Users are better at identifying fully GPT-written abstracts with the accuracy of 43.81\% and 66.13\%, while they perform worse with GPT-polished abstracts with the accuracy of 37.5\% and 50.88\%. 

\noindent$\bullet~$ Users' self-claimed expertise appears to very slightly affect their capability to identify human-written and GPT-generated abstracts. For example, participants with ``No familiarity'' with papers performed worse than the others. However, most of the differences are \textit{not} statistically significant. 
    
\noindent$\bullet~$ Users become slightly better at identifying GPT-generated content when they are provided with references. However, the accuracy is still too low for reliable identification. The largest improvement is witnessed in HSS with an accuracy improvement of 21.5\%.

%% file: tex/checkgpt.tex
\section{\checkgpt: An Accurate Detector for ChatGPT-generated Academic Writing}\label{sec:checkgpt}

\subsection{The System Model and Assumptions}\label{subsec:model}


Our objective is to build a classifier, \checkgpt, to determine whether a given text snippet is generated by ChatGPT. We denote our classifier as $\mathcal{H}$, and the classification problem can be formulated as:

\vspace{-3mm}
\begin{equation}
\hat{y}=\mathcal{H}_{\theta}(\mathbf{s})
\end{equation}
\begin{equation}
{\argmin}_\theta\:\mathcal{L}(y, \hat{y})
\end{equation}
where $\mathbf{s}$ is an unstructured text snippet (i.e., paper abstract). Given $\mathbf{s}$, $\mathcal{H}(\mathbf{s})$ generates the probability distribution $\hat{y}$ considering label space $\{\text{`h'}, \text{`g'}\}$, where `h' indicates human-written text and `g' indicates ChatGPT-generated text. The goal is to find an optimal set of parameter $\theta$ for $\mathcal{H}$ to minimize the loss function $\mathcal{L}$ measuring the distance between prediction $\hat{y}$ and observation ${y}$.

We consider a \textit{black-box defender}, who only has access to the observed LLM outputs, and does not have insider knowledge of the LLM that generates these samples (weights, structures, and gradients). This assumption is practical, considering OpenAI has not open-sourced its LLMs since the GPT-3.5. 

The primary goal of this work is to detect \textit{raw} ChatGPT-generated text, i.e., the unaltered output from ChatGPT. In the community, there have been discussions on the boundary between GPT-generated text and Human-generated text in the context of partial paraphrasing \cite{acl2023policy}, where the boundary can become murky if a piece of GPT-generated text is further edited by a human editor. Therefore, to avoid the controversy of data labeling, 
we mainly focus on the clean boundary between purely human-generated text and raw ChatGPT output in problem setting\footnote{We will relax this assumption in Section \ref{sec:robustness} and evaluate \checkgpt's robustness against \textit{post facto} human interventions}.


Based on the application of the task, we further assume the following demands in addition to the defender's goals: (1) \textit{Moderate Data Availability}. We do not assume the defender's privileged usage of ChatGPT. Therefore, the training samples are collected strictly following OpenAI's policy. With the efficiency of regular queries, an ordinary user usually can not collect massive amounts (e.g., tens of millions to billions) of data. 
(2) \textit{Affordability.} We do not assume the defender's access to excessive computing power, which is only affordable to large organizations in real world. We aim to offer a lightweight solution that smaller entities could conveniently obtain and deploy in a daily operational environment. And (3) \textit{Privacy-Preserving Local Deployment.} The end users may not agree to share their data with the detector providers due to privacy, intellectual property, or policy concerns (e.g., manuscripts being reviewed). Therefore, the detector should have the potential to be transferred to a new domain with a small amount of target-doman data.

\subsection{Linguistic \& Semantic-Based Detection}\label{subsec:baseline}
{
We first explore a non-deep learning method using hand-crafted linguistic \& semantic features as our baseline approach. The raw texts are transformed into vector representations using the NELA (News Landscape) features \cite{nela}. The NELA features were initially proposed to analyze news articles. They comprise six linguistic and semantic features: styles, complexity, biases, affects, morals, and events. We adopt Gradient Boosting (GB) decision trees to distinguish between human-written and GPT-generated abstracts. For each task and discipline, the models are trained with 50,000 human-written and GPT-generated samples, respectively, and tested with 10,000 samples from each class, i.e., an 80/20 train-test split ratio. 

The classification performance of the baseline models is shown in Table \ref{tab:baseline}. We have the following observations: (1) For Task 1, GB performs well in distinguishing GPT-written abstracts in CS and physics. (2) For Task 2, the F1-scores drop to about 0.9, indicating an increased difficulty. (4) The GPT-polished abstracts in Task 3 are more challenging to detect. The performance decreases significantly to <0.8 for all the three disciplines. 

We inspect the feature importance within GB model to analyze the difference between the GPT-generated and human-written abstracts. In general, we find that ChatGPT composes differently in several characteristics: (1) ChatGPT tends to compose longer sentences and use longer words during writing; (2) ChatGPT habitually uses more gerunds as adverbials; (3) ChatGPT uses more determiners, more gerunds, and fewer personal pronouns than human writers.
More details can be found in Appendix~\ref{apdx:nelagb}.

While the baseline approach lacks capabilities in some tasks, it still significantly outperforms the human evaluators tested in Section~\ref{sec:survey}, and the existing detecting tools in 
Table~\ref{tab:otherdetectors}. It showcases the different habitual patterns of GPT-generated and human-written abstracts. Prominently, ChatGPT exhibits a tendency of increased complexity and a higher frequency of third-person singulars and gerunds. While these habitual patterns identified provide valuable insights, they alone cannot guarantee reliable and generalizable classification, as shown by the model's insufficiency on Task 2 and 3. This motivates us to adopt DNN-based detectors, which are more capable of capturing subtler and deeper semantic and/or linguistic patterns that the hand-crafted NELA features may overlook.

\begin{table}[t]
	\caption{The baseline approach: classification F1-score for NELA classifier.}\label{tab:baseline}\vspace{-3mm}
    \centering
    {\small 
	\begin{tabular}{ccc|ccc|ccc}
     \multicolumn{3}{c|}{T1. GPT-WRI} &\multicolumn{3}{c|}{T2. GPT-CPL}  &\multicolumn{3}{c}{T3. GPT-POL}\\ \hline
        CS & PHX & HSS & CS & PHX & HSS & CS & PHX & HSS\\\hline
0.965	&	0.980	&	0.963	&	
0.901	&	0.918	&	0.896	&	
0.774	&	0.794	&	0.798	\\
        \end{tabular}
        }\vspace{-3mm}
\end{table}

}

\subsection{The CheckGPT Framework}
\noindent\textbf{{\em Preliminaries.}} The Bidirectional Encoder Representations from Transformers (BERT) \cite{devlin2018bert} family, including but not limited to BERT itself and RoBERTa, have shown extraordinary capabilities in a wide range of NLP tasks. RoBERTa (Robustly Optimized BERT approach) \cite{liu2019roberta}, is the state-of-the-art member of this family built upon BERT by Meta. Models like RoBERTa are pre-trained on a massive corpus from diverse disciplines. Such extensive training allows them to capture and represent various linguistic patterns, syntactic structures, and semantic relationships in the texts. Its tokenization and encoding enable the transformation of raw data into effective representations, which can be used for downstream tasks. In this work, we utilize the pre-trained RoBERTa to preprocess the text data. The pre-training of the RoBERTa utilizes a masked language modeling (MLM) objective, which can be formalized as:

\begin{equation}
\mathcal{L}_{\text{MLM}}=-\mathbb{E}_{\mathbf{s} \sim \mathcal{D}_{s}}{\log P(m | \mathbf{s})}
\end{equation}
\noindent where $\mathcal{D}_{s}$ is the corpus, $\mathbf{s}$ denotes an input sequence, and $m$ is a masked token. The representations extracted by RoBERTa serve as the features for our downstream classification head. Long-Short-Term Memory (LSTM) networks \cite{hochreiter1997long}, a variant of Recurrent Neural Network (RNN) which incorporates gating mechanism to effectively retain information, can improves feature extraction over long sequences. In this work, we adopt LSTM to build the classification head to aggregate the sequences of RoBERTa encoded tokens, and to fine-tune the final representation for task-specific prediction.

\begin{figure}[!t]
\centering
\includegraphics[width=\columnwidth]{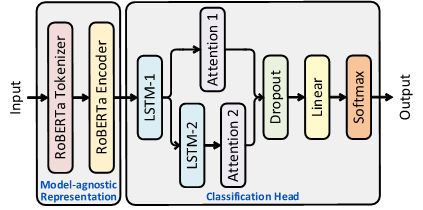} 
\caption{The architecture of the \checkgpt~model.} 
\label{fig:checkgpt}\vspace{-3mm}
\end{figure}

\noindent\textbf{{\em Representation.}} 
\checkgpt~ framework consists of two stages: representation and classification, as shown in Figure \ref{fig:checkgpt}.
For representation, \checkgpt~proposes a model-agnostic design for text encoding where any pre-trained representation models (e.g., other variants of BERT) are directly employed as is. This design achieves higher affordability, upgradability, and flexibility since 
(1) adopting a sophisticated pre-trained embedding model can saves tremendous efforts and computations compared with training an embedding layer from scratch, which is often beyond the capability of regular users or organizations;
(2) a plugin design allows future upgrades by seamlessly accommodating new representation models; 
and 
(3) the lightweight classification head is easier to fine-tune when new data or domains are added.
In \checkgpt, the representation stage is completed using the tokenizer and encoders of RoBERTa-large\footnote{https://github.com/facebookresearch/fairseq/tree/main/examples/roberta}. For tokenization, the pre-trained RoBERTa-large enforces a limit of 512 tokens. The tokenization can be formalized as:
\begin{equation}
\mathbf{X} = \text{Tokenizer}(\mathbf{s}) = \{x_{i}\}_{i=1}^{n}
\end{equation}
\noindent where $\mathbf{X}$ denotes a sequence of length $n$ consisting of individual tokens $x_{i}$, and $\text{Tokenizer}$ refers to the Byte-level Pairing Encoding (BPE) utilized by RoBERTa.

For the encoder, the RoBERTa uses embeddings of size 1024 to represent each token. In this way, the tokenized texts are encoded into contextualized representations with a shape of $n \times 1024$. The encoding can be formalized as:
\begin{equation}
\mathbf{E} = \text{Encoder}(\mathbf{X}) = \{e_{i}\}_{i=1}^{n},\;\;   e_{i} \in \mathbb{R}^{1024} 
\end{equation}
\noindent where $\mathbf{E}$ denotes a sequence consisting of individual embedding $e_{i}$, and $\text{Encoder}$ refers to the transformer encoder utilized in RoBERTa.

\noindent\textbf{{\em Classification.}} The derived embeddings are fed into the LSTM-based classification stage. The classification head consists of two LSTMs (with embedding size of 256), and each has a following attention layer \cite{baziotis-etal-2018-ntua-slp-semeval} to aggregate the entire sequence into a single vector. The outputs from the two LSTMs modules are later concatenated and passed through a dropout layer with a rate of $p=0.5$ for overfitting prevention, and a dense layer (FC) for final task-specific fine-tuning. 
The softmaxed output indicates the conditional probability of the two classes: ``GPT-generated'' ($y_{g}$) or ``Human-generated'' ($y_{h}$). The functions of our LSTM classifier $f_{\theta}(\mathbf{E})$ are as follows:
\begin{equation}
\begin{aligned}
h_{1} = \text{LSTM}_{1}(\mathbf{E}),\;\;r_{1} = \text{ATTN}_{1}(h_{1}) \\
h_{2} = \text{LSTM}_{2}(h_{1}),\;\;r_{2} = \text{ATTN}_{2}(h_{2}) \\
(\hat{y}_{g}, \hat{y}_{h}) = \text{Softmax}(\text{FC}(\text{Dropout}(r_{1} \oplus r_{2})))
\end{aligned}
\end{equation}
\noindent\textbf{{\em Model Training.}} The classifier $f_{\theta}$ with parameter $\theta$ is optimized independently with the RoBERTa frozen (as is) during the training. We adopt an AdamW optimizer \cite{loshchilovdecoupled}, a CosineAnnealing learning rate scheduler \cite{loshchilovsgdr}, and a gradient scaler for efficient mixed-precision training \cite{micikeviciusmixed}. 
Given the model's predicted probabilities $\hat{y}=(\hat{y}_{h}, \hat{y}_{c})$ and one-hot encoded ground truth $y=(y_{h}, y_{c})$
, the Binary Cross Entropy (BCE) loss of each training sample is defined as: 
\begin{equation}
\mathcal{L}(\theta)=-\left[y_{c} \log(\hat{y}_{c}) + y_h \log(\hat{y}_{h})\right]
\end{equation}

\noindent\textbf{{\em Design Choices and Discussions.}} One alternative approach is directly applying RoBERTa by adding a RobertaClassificationHead \cite{robertahuggingface}. However, experiments show that \checkgpt~incur a higher accuracy, which can be attributed to LSTM's capability to track the sequential dependencies over long periods in the text sequences \cite{yin2017comparative}. Please refer to Section~\ref{sec:baselineperf} for details of the ablation study.


Another alternative approach is to include the entire pre-trained language representation model \cite{ott2019fairseq, robertacustom} for task-specific fine-tuning. As shown in Table \ref{tab:opensourcedetector}, \checkgpt~ consistently outperforms the approaches of fine-tuning the entire language representation models (BERT, DistillBERT, RoBERTa, and GPT-2) on GPABench2.

Comparing with both alternative approaches in framework design, \checkgpt~ offers the following advantages: 
(1) \textbf{Efficiency}: \checkgpt~can significantly reduce the parameters to save both time and computing resources. Given the parameters of language models ranging from 66M (DistilledBERT \cite{sanh2019distilbert}) to 355M (RoBERTa-large, GPT-2-medium) and 1750M (GPT-3 \cite{brown2020language}), our model only maintain 4M parameters and achieve satisfactory accuracy. The drop in model size also reduces the risks of over-fitting and catastrophic forgetting \cite{mosbach2021stability, kirkpatrick2017overcoming}, especially with small datasets \cite{uchendu2020authorship, bakhtin2019real}.
(2) \textbf{Applicability}: Our framework is model-agnostic thus accepts various representation approaches (e.g., BERT\cite{devlin2018bert}, BART \cite{lewis2019bart}), making it a lightweight and universal detector, as detailed in Section~\ref{sec:baselineperf}.
This feature can especially valuable considering deployment and customization in real world.
(3) \textbf{Versatility}: By freezing RoBERTa's well-crafted parameters trained on a broad range of domains, we retain the framework's generalizability to a great extent to use \checkgpt~ in different subjects, which is challenging for fine-tuned RoBERTa \cite{wang2023m4}. More details are presented in Section~\ref{sec:transfer}.


%% file: tex/evaluation.tex
\section{Experiments}\label{sec:exp}

\subsection{Settings and Metrics}

We implement \checkgpt~with PyTorch 1.13.1 in Python 3.9.1 on Ubuntu 22.04. The pre-trained RoBERTa is adopted from \cite{robertahuggingface}. All the experiments were conducted on an Nvidia 2080Ti GPU and an Intel i9-9900k CPU. We use GPABench2 for most of the experiments. \checkgpt~is trained with an initial learning rate of 2e-4, a batch size of 256, and an early-stop strategy to finish training when the validation loss stays flat for a predefined number of epochs. 

When we consider \checkgpt~as GPT-generated content detector, the \textit{true positive} rate ($TPR=\frac{TP}{TP+FN}$) is the proportion of correctly detected GPT-generated abstracts out of all GPT-generated abstracts, i.e., the accuracy in classifying GPT-generated text. The \textit{true negative} rate ($TNR=\frac{TN}{TN+FP}$), is the proportion of correctly identified human-written abstracts out of all human-written abstracts, i.e. the accuracy in classifying human-written text. The overall accuracy is defined as the proportion of correctly classified samples over all the testing samples: $Acc=\frac{TP+TN}{TP+FP+TN+FN}$. 




\subsection{Task- and Discipline-specific Classifiers}\label{sec:baselineperf}

\begin{table}[t]
	\caption{\checkgpt's performance (in \%) for each task, discipline, and prompt: TPR, TNR, accuracy (Acc).}\label{tab:checkgpt}
    \setlength{\tabcolsep}{0.3em}
    \centering
    \vspace{-2mm}
    {\small
	\begin{tabular}{c|ccc|ccc|ccc}
        \hline
       & \multicolumn{3}{c|}{T1. GPT-WRI} &\multicolumn{3}{c|}{T2. GPT-CPL}  &\multicolumn{3}{c}{T3. GPT-POL}\\\cline{2-10}
         & CS & PHX & HSS & CS & PHX & HSS & CS & PHX & HSS\\\hline
           \multicolumn{10}{c}{Prompt 1} \\\hline
TPR & 99.95 & 100.0 & 99.94 & 99.58 & 99.50 & 99.48 & 99.23 & 99.31 & 99.22 \\
TNR & 99.97 & 100.0 & 100.0 & 99.70 & 99.57 & 99.67 & 99.33 & 99.40 & 99.39 \\
ACC & 99.96 & 100.0 & 99.97 & 99.64 & 99.54 & 99.58 & 99.28 & 99.36 & 99.31 \\
\hline

\multicolumn{10}{c}{Prompt 2} \\\hline
TPR & 99.99 & 100.0 & 100.0 & 99.63 & 99.50 & 99.65 & 99.23 & 99.50 & 99.32 \\
TNR & 99.99 & 100.0 & 99.98 & 99.75 & 99.63 & 99.68 & 99.33 & 99.51 & 99.40 \\
ACC & 99.99 & 100.0 & 99.99 & 99.69 & 99.57 & 99.67 & 99.28 & 99.51 & 99.36 \\
\hline

\multicolumn{10}{c}{Prompt 3} \\\hline
TPR & 99.97 & 99.99 & 99.96 & 99.78 & 99.68 & 99.67 & 99.33 & 99.35 & 99.46 \\
TNR & 100.0 & 100.0 & 99.98 & 99.76 & 99.79 & 99.71 & 99.34 & 99.68 & 99.36 \\
ACC & 99.98 & 100.0 & 99.97 & 99.77 & 99.74 & 99.69 & 99.34 & 99.52 & 99.41 \\
\hline

\multicolumn{10}{c}{Prompt 4} \\\hline
TPR & 100.0 & 99.99 & 99.96 & 99.76 & 99.71 & 99.67 & 99.41 & 99.61 & 99.47 \\
TNR & 99.99 & 100.0 & 99.99 & 99.77 & 99.78 & 99.88 & 99.60 & 99.69 & 99.50 \\
ACC & 100.0 & 100.0 & 99.98 & 99.77 & 99.75 & 99.78 & 99.51 & 99.65 & 99.49 \\
\hline

        \end{tabular}} \vspace{-4mm}



\end{table}

We first evaluate \checkgpt~at the finest granularity: one classifier for each discipline, task, and prompt combination. We use an 80\%-20\% train-test split on the main GPABench2 dataset: 80K samples (40K each of GPT and HUM) for training and 20K for testing. 
Training takes an average of 120s per epoch, while testing takes about 0.03s per sample. We report the classification accuracy in Table~\ref{tab:checkgpt}. 


\checkgpt~achieves very high accuracy in all cases. The detection accuracy for Task 1 (abstracts entirely written by ChatGPT) is higher than 99.9\% across all disciplines/prompts.  
Task 2, where only the second halves of the abstracts are checked, has slightly lower accuracy, which is explained by shorter text lengths and better writing by ChatGPT given more seed data.
The accuracy of Task 3, which is most challenging for the open-source and commercial detectors (Sec. \ref{sec:benchmark}), is between 99.28\% and 99.65\%. 

\begin{figure}[!t]
\centering
\includegraphics[width=0.98\columnwidth]{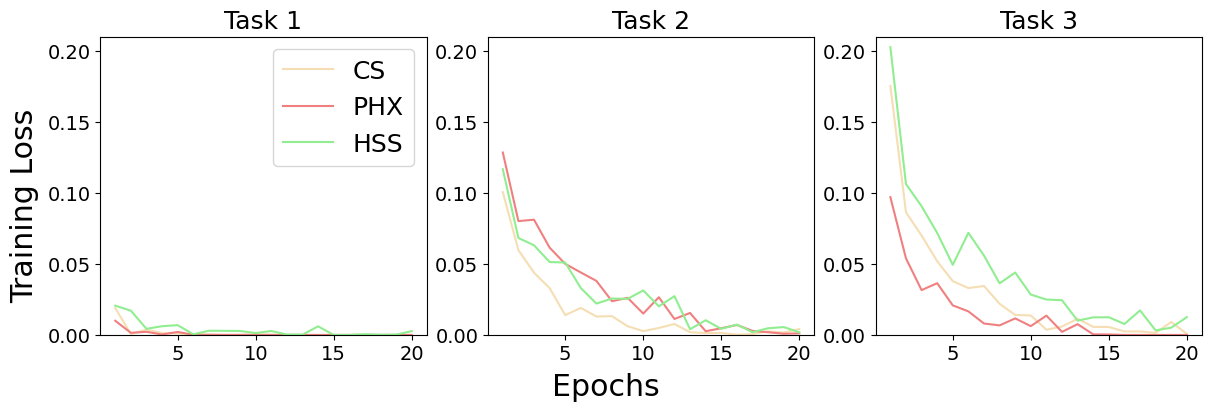}\vspace{-3mm}
\caption{Training losses of the task-specific and discipline-specific classifiers.}\vspace{-2mm}
\label{fig:trainloss} \vspace{-2mm}
\end{figure}

Figure \ref{fig:trainloss} shows the training losses (Prompt 1). Task 1 models rapidly grasp simple features like lexical characteristics, while Tasks 2 and 3 are more difficult. In most cases, HSS is more challenging, which implies that ChatGPT does a better job mimicking human-written style in these topics. Task 2 is the outlier, where the samples in PHX are significantly shorter and thus harder to distinguish.

\begin{figure}[!t]
\centering
\includegraphics[width=0.98\columnwidth]{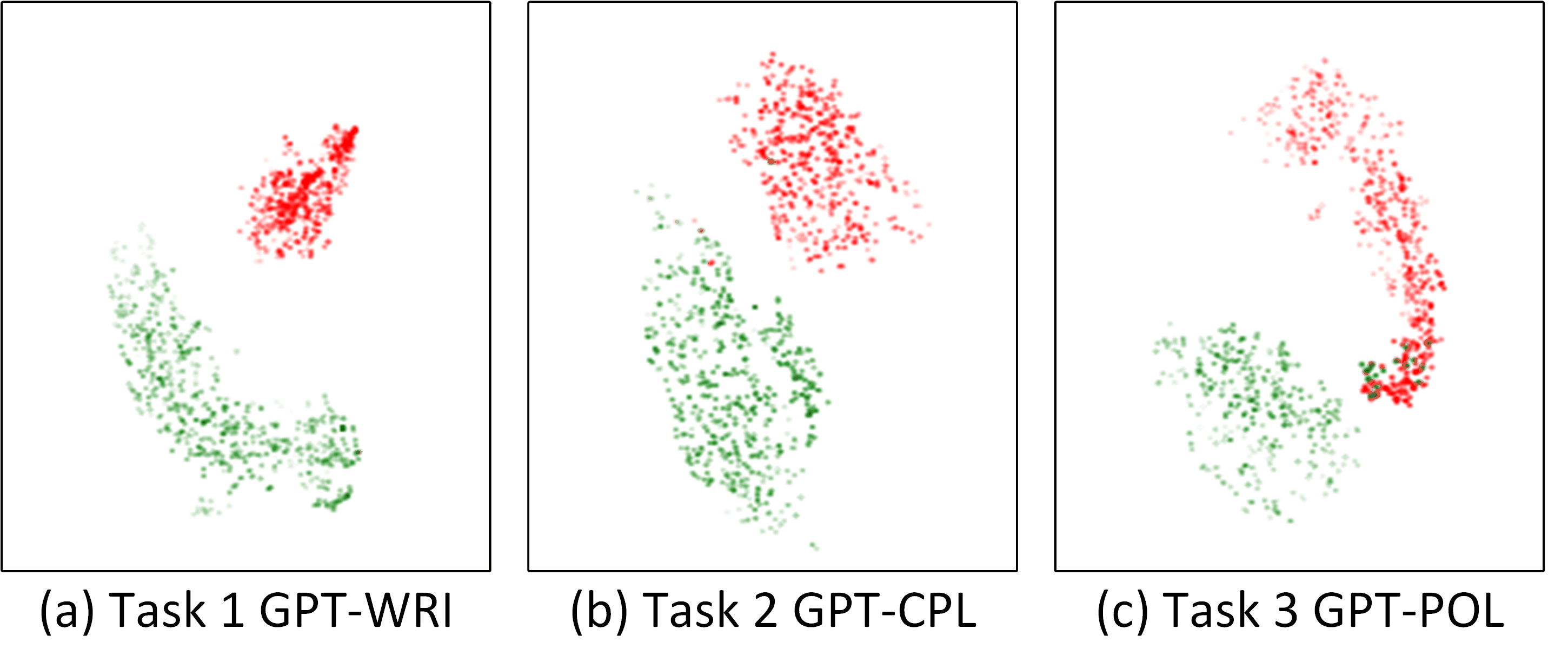}\vspace{-3mm}
\caption{Feature space distribution of human-written (green) and GPT-generated (red) abstracts.}
\label{fig:visualize}
\end{figure}

We randomly select 2,000 CS abstracts from each task/label, and then use t-Distributed Stochastic Neighbor Embedding (t-SNE) \cite{van2008visualizing} to map the 1024-D feature vectors from the last dense layer of the BiLSTM module into a 3-D space, as shown in Figure~\ref{fig:visualize}. The figure shows that: (a) GPT-written abstracts form a dense cluster (consistent vocabulary, writing style, and semantic features), which is different from the varied distribution of human-written samples. (b) GPT-completed abstracts (Task 2) are significantly more diverse than the GPT-written ones. While their representations are closer to the human-written samples, a distinct gap remains. (c) GPT-polished samples are scattered and intertwined with human-written samples, demonstrating the challenges in Task 3.

\noindent\textbf{\em Ablation study.}  We compare the current design of \checkgpt~with several alternatives. 
We first keep the classification head in \checkgpt~and replace the representation module with GLoVe6B-100d \cite{pennington2014glove} or pre-trained BERT. As shown in Table \ref{tab:ablation}, there is a slight performance drop. We then keep the representation module and replace the classifier with: the default classification head for RoBERTa (RCH) \citet{robertahuggingface}, and its variant with global average pooling (MLP-Pool; \citealp{lin2013network, keraspool}); an AlexNet-like CNN \cite{krizhevsky2012imagenet} with five convolutional layers, and (4) a basic BiLSTM without attention.  As shown in Table \ref{tab:ablation}, \checkgpt~achieves the best accuracy. 



\begin{table}[t]
\caption{Comparison with other design choices.} \label{tab:ablation}
\setlength{\tabcolsep}{0.3em}
\centering
\vspace{-3mm}
{\small
\begin{tabular}{c|c|ccc}
\hline
\multirow{2}{*}{Model} & Para & \multicolumn{3}{c}{Acc(\%)} \\ \cline{3-5}
 & Size & Task 1 & Task 2 & Task 3 \\\hline
 GLoVe + \checkgpt~classifier & - & 99.77 & 98.34 & 95.90 \\
 BERT + \checkgpt~classifier & - & 99.90 & 99.28 & 97.81 \\\hline
\checkgpt~representation + RCH & 1.05M & 99.80 & 97.70 & 94.08 \\
\checkgpt~representation + MLP-Pool & 1.05M & 99.87 & 98.62 & 95.93 \\
\checkgpt~representation + CNN & 3.33M & 99.80 & 98.47 & 96.49 \\
\checkgpt~rep. + BiLSTM w/o attention & 4.21M & 99.91 & 99.54 & 98.92 \\\hline
\checkgpt & 4.21M & 99.98 & 99.72 & 99.39 \\
\hline
\end{tabular}} \vspace{-3mm}
\end{table}



\begin{figure}[!t]
\centering
\includegraphics[width=\columnwidth]{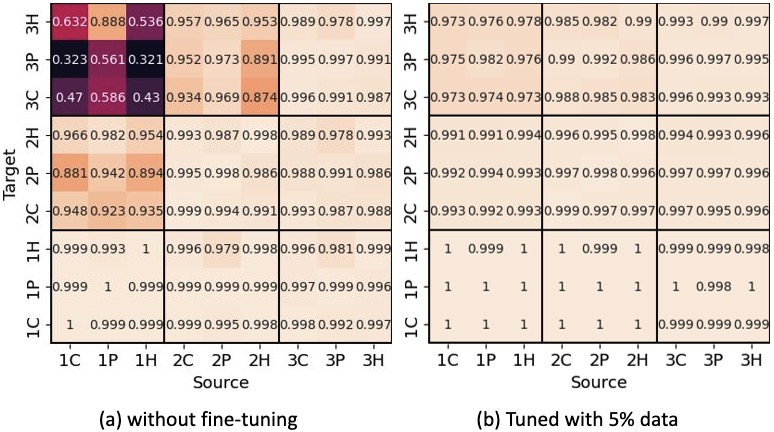}\vspace{-2mm}
\caption{\checkgpt's transferability across disciplines and tasks: (a) without fine-tuning, (b): tuned with 5\% data from the train set. 1C: Task 1 GPT-WRI+CS; 2P: Task 2 GPT-CPL+PHX; 3H: GPT-POL+HSS. }
\label{fig:transfer} \vspace{-2mm}
\end{figure}

\subsection{Transferability across Tasks, Disciplines, and Prompts}\label{sec:transfer}
We evaluate \checkgpt's capability of cross-prompt, cross-task, and cross-disciplinary generalization. First, we train nine cross-prompt models (one model for each task and discipline as shown in Table \ref{tab:unified} (a)) to evaluate testing samples from \textbf{other} tasks and disciplines, \textit{without model fine-tuning}. In Figure \ref{fig:transfer} (a), each value demonstrates the F1-score using the model from the task/discipline denoted on the x-axis to test samples from the task/discipline on the y-axis. From the figure, we observe the following:

\noindent$\bullet~$ \checkgpt~is adaptable \textit{across disciplines}. \checkgpt~achieves $\geq$0.978 F1-score on \textit{cross-discipline} data from the \textit{same task}.

\noindent$\bullet~$ \checkgpt~is less adaptable \textit{across tasks}. In particular, the models trained in Task 1 demonstrate low performance with testing samples from the other tasks, while the models from Task 2 are also less capable of handling Task 3 (GPT-POL) data.

\noindent$\bullet~$ The models trained in Task 3 demonstrate solid performance with testing samples from Tasks 1 and 2. It implies that Task 3 could be the most difficult, and the models have learned subtle but inherent features of AIGC.


We then fine-tune the last linear layer of each model with the data in the target domain. As shown in Figure \ref{fig:transfer} (b), tuning with as few as 5\% of data (2K samples) increases the classification F1-score to 0.97+ in all cases, while the distribution patterns of the F-1 scores remain similar to Figure \ref{fig:transfer}(a). 


\begin{table}[t]
	\caption{TPR and TNR (in \%) of the unified classifiers.}\label{tab:unified}
    \setlength{\tabcolsep}{0.39em}
    \centering
    \vspace{-3mm}
    {\small
	\begin{tabular}{c|ccc|ccc|ccc}
    \hline
        & \multicolumn{3}{c|}{T1. GPT-WRI} &\multicolumn{3}{c|}{T2. GPT-CPL}  &\multicolumn{3}{c}{T3. GPT-POL}\\ \cline{2-10}
         & CS & PHX & HSS & CS & PHX & HSS & CS & PHX & HSS\\ \hline
         \multicolumn{10}{c}{(a) Cross-prompt Classifiers}\\\hline
TPR & 99.99 & 99.99 & 99.99 & 99.84 & 99.87 & 99.78 & 99.74 & 99.75 & 99.74 \\
TNR & 99.97 & 100.0 & 99.97 & 99.52 & 99.43 & 99.51 & 98.87 & 99.31 & 98.90 \\
\hline

 \multicolumn{10}{c}{(b) Cross-prompt Cross-disciplinary Classifiers}\\\hline
TPR & 99.99 & 99.99 & 99.99 & 99.82 & 99.72 & 99.72 & 99.62 & 99.73 & 99.62 \\
TNR & 99.99 & 100.0 & 99.99 & 99.66 & 99.67 & 99.72 & 99.10 & 99.58 & 99.24 \\
\hline

         \multicolumn{10}{c}{(c) Cross-prompt \& -task \& -disciplinary Classifier}\\\hline
TPR & 100.0 & 100.0 & 100.0 & 99.81 & 99.77 & 99.80 & 99.58 & 99.76 & 99.65 \\
TNR & 99.16 & 99.46 & 99.27 & 99.47 & 99.45 & 99.55 & 99.16 & 99.46 & 99.27 \\
\hline
        \end{tabular}} 
\end{table}


\noindent\textbf{\em Prompt Transferability.} To assess \checkgpt's generalizability over domain shifts caused by ChatGPT prompts, we train it with data from 3 prompts and test it with samples from the fourth prompt. As shown in Table~\ref{tab:3to1},  \checkgpt~is highly transferable across prompts. Notably, testing accuracy for Prompt 1 is usually the lowest, which is explained by the fact that Prompt provided minimal context to ChatGPT so that the outputs are most diverse. 


\begin{table}[t]
	\caption{TPR and TNR (in \%) of cross-prompt testing.}\label{tab:3to1}
    \setlength{\tabcolsep}{0.3em}
    \centering
    \vspace{-3mm}
    {\small
	\begin{tabular}{c|ccc|ccc|ccc}
    \hline
        & \multicolumn{3}{c|}{T1. GPT-WRI} &\multicolumn{3}{c|}{T2. GPT-CPL}  &\multicolumn{3}{c}{T3. GPT-POL}\\ \cline{2-10}
         & CS & PHX & HSS & CS & PHX & HSS & CS & PHX & HSS\\\hline
         \multicolumn{10}{c}{(a) Train with Prompts 2, 3, 4; test with Prompt 1}\\\hline
TPR & 99.73 & 99.91 & 99.63 & 99.43 & 99.31 & 99.54 & 97.73 & 98.47 & 98.00 \\
TNR & 99.89 & 99.99 & 99.95 & 99.35 & 99.42 & 99.19 & 98.54 & 99.00 & 98.99 \\
\hline

\multicolumn{10}{c}{(b) Train with Prompts 1, 3, 4; test with Prompt 2}\\\hline
TPR & 99.46 & 99.83 & 99.59 & 99.63 & 99.50 & 99.55 & 98.82 & 99.18 & 99.59 \\
TNR & 99.89 & 99.81 & 99.85 & 99.14 & 99.40 & 99.20 & 99.23 & 99.39 & 98.87 \\
\hline

\multicolumn{10}{c}{(c) Train with Prompts 1, 2, 4; test with Prompt 3}\\\hline
TPR & 99.99 & 99.99 & 99.97 & 99.31 & 99.51 & 99.60 & 99.34 & 99.75 & 99.68 \\
TNR & 99.87 & 99.92 & 99.89 & 99.49 & 99.46 & 99.29 & 98.86 & 98.98 & 98.80 \\
\hline

\multicolumn{10}{c}{(d) Train with Prompts 1, 2, 3; test with Prompt 4}\\\hline
TPR & 99.98 & 99.95 & 99.95 & 99.67 & 99.39 & 99.51 & 99.75 & 99.63 & 99.79 \\
TNR & 99.79 & 99.91 & 99.82 & 99.06 & 99.41 & 99.35 & 98.47 & 98.99 & 98.79 \\
\hline
\end{tabular}}\vspace{-2mm}
\end{table}


\noindent\textbf{\em The Unified Classifiers.} We evaluate \checkgpt~by sampling data from all prompts (train with 80K GPT samples for each task/discipline) and show the classification accuracy in Table~\ref{tab:unified}~(a). We then combine data across disciplines (Table~\ref{tab:unified} (b)) and further across all tasks 
(Table~\ref{tab:unified} (c)). In summary, unified training slightly improves TPR, especially for difficult tasks, e.g., GPT-POL in CS.

\noindent\textbf{\em The Multi-Task Classifier.} Finally, we investigate whether \checkgpt~is able to further distinguish the difference among the three tasks. To do this, we turn \checkgpt~into a unified, 4-ary classifier to distinguish HUM, GPT-WRI, GPT-CPL, and GPT-POL abstracts. \checkgpt~achieved a 98.51\% accuracy, with 98.75\%, 99.39\%, 97.69\%, and 98.24\% for HUM, GPT-WRI, GPT-CPL, and GPT-POL, respectively. \checkgpt's accuracy drops slightly due to the challenges of multi-label classification, but it can still catch the subtle differences in three tasks well.



\subsection{Transferability to New Domains}\label{subsec:newdomain}

\textbf{\em Other academic writing purposes.} While our GPABench2 focuses on abstracts of research papers, ChatGPT can be used for other writing purposes, e.g., other sections of papers. While paper abstracts are almost always publicly available, the full papers are often restricted by copyright. Web scraping is not allowed even for open-access papers. Thus, we do not attempt to include human-written full papers in GPABench2. Instead, we provide ChatGPT with titles and ask it to write introduction, background, and conclusion sections for the title (similar to Task 1).
For each discipline and section, we collect 10,000 samples using the titles from GPABench2 (90,000 in total). We apply the cross-prompt cross-discipline \checkgpt~(Sec. \ref{sec:transfer}) on the new sections. As shown in Table \ref{tab:otherpurpose}, \checkgpt~maintains 99.9\% accuracy for all the sections. This shows the high potential of \checkgpt~for general academic writing.


\begin{table}[t]
	\caption{TPR (in \%) of other writing purposes.}\label{tab:otherpurpose}
    \setlength{\tabcolsep}{0.3em}
    \centering
    \vspace{-3mm}
    {
	\begin{tabular}{c|ccc|ccc|ccc}
 \hline
        & \multicolumn{3}{c|}{Introduction} &\multicolumn{3}{c|}{Background}  &\multicolumn{3}{c}{Conclusion}\\ \cline{2-10}
         & CS & PHX & HSS & CS & PHX & HSS & CS & PHX & HSS\\ \hline
TPR & 99.99 & 100.0 & 100.0 & 99.99 & 100.0 & 99.98 & 100.0 & 100.0 & 100.0 \\
\hline
        \end{tabular}}
\end{table}

\noindent
\textbf{\em Classic NLP Datasets.} We evaluate \checkgpt~with three NLP datasets: Wiki Abstracts (1,500 random samples from \citealp{brummer2016dbpedia}), ASAP Essays \cite{hewlettessaydataset}, and BBC News \cite{greene06icml}. In ASAP Essays, we selected two different tasks: ``letters stating opinions on computers'' (Essay-C), and ``stories about patience'' (Essay-P). We adopted the original instructions in \citet{hewlettessaydataset} for Task 1 and designed the prompts for the other tasks and datasets (see Appendix~\ref{apdx:newdomain} for details).
We apply the cross-prompt cross-discipline \checkgpt~classifiers (Sec. \ref{sec:transfer}) on the new domains. As shown in Table \ref{tab:otherdataset}, \checkgpt~shows solid performance, especially on objective, structural, or argumentative writing like news and opinions. When the last layer of \checkgpt~is tuned with 100 samples (50 for each label) from each domain, it achieves significantly higher accuracy. 

\noindent
\textbf{\em SOTA ChatGPT Datasets in the Literature.}
We further evaluate \checkgpt~on five ChatGPT datasets in the literature that cover 15 domains. 
We adopted the following datasets: student essays in ArguGPT \cite{liu2023argugpt}; Finance, Medicine, OpenQA, Reddit, and Wikipedia in HC3 \cite{guo2023close}; arXiv papers, PeerRead reviews, Reddit, wikiHow manuals, and Wikipedia in M4 \cite{wang2023m4}; news articles in  MULTITuDE \cite{macko2023multitude}; and MGTBench \cite{he2023mgtbench} that covers short Q\&A from NarrativeQA \cite{kovcisky2018narrativeqa}, SQuAD1.0 \cite{rajpurkar2016squad} and TruthfulQA \cite{lin2022truthfulqa}. 

We evaluate \checkgpt~with three experiments: (1) directly validate \checkgpt~in without any fine-tuning, (2) fine-tuning the last layer of the classification head with 150 samples, and (3) fine-tuning the whole classification head with the entire training set. We also compare \checkgpt~with fine-tuning the entire language models as reported in these papers. 
As shown in Table~\ref{tab:otherdataset}, \checkgpt~reaches an F1-score of >0.95 for 9 out of the 15 domains in direct validations without finetuning, and reaches >0.9 for two more domains. Unsatisfactory performance are shown on wikiHow and the short Q\&A in MGTBench. Unlike all the writing tasks, the text in wikiHow is highly informal with many imperatives as tips or advice. For the MGTBench, most of the answers consist of a single sentence, and many of them even come with one or two words. For these four domains, the gaps are too large for \checkgpt~to transfer effectively. 
Notably, the fully-tuned \checkgpt~outperforms fine-tuned language models in almost all the domains except M4-Wikipedia, where \checkgpt~is 0.012 lower than RoBERTa. Note full tuning of \checkgpt~takes approximately 120s per epoch, which is considerably less expensive than tuning RoBERTa (1049s per epoch) or DistillBERT (548s per epoch). 

\begin{table}[t]
	\caption{F1-score of evaluating \checkgpt~on SOTA datasets in different domains. Val.: Direct Validation. FT-L: Fine-tuning the last layer. FT-A: Fine-tuning all the layers in the \checkgpt~classifier. All SOTA detectors were trained/tuned with data in the target domain.}\label{tab:otherdataset}
    \setlength{\tabcolsep}{0.39em}
    \vspace{-3mm}
    {
	\begin{tabular}{c|ccc|cc}
 \hline
   \multirow{2}{*}{Dataset} & \multicolumn{3}{c|}{\checkgpt} & \multicolumn{2}{c}{SOTA} \\ \cline{2-6}
   & Val. & FT-L & FT-A & Model & F1-score \\\hline
     ArguGPT-WECCL & 1.000 & 1.000 & 1.000 & RoBERTa & 0.994 \\ \hline
     HC3-Finance & 0.957 & 0.971 & 0.998 & RoBERTa & 0.993 \\
     HC3-Medicine & 0.985 & 0.985 & 0.998 & RoBERTa & 0.995 \\ 
     HC3-OpenQA & 0.936 & 0.948 & 0.999 & RoBERTa & 0.986 \\
     HC3-Reddit & 0.965 & 0.973 & 1.000 & RoBERTa & 1.000 \\
     HC3-Wiki & 0.945 & 0.971 & 0.991 & - & - \\ \hline
     M4-arXiv & 0.995 & 0.999 & 1.000 & RoBERTa & 1.000 \\
     M4-PeerRead & 0.962 & 0.968 & 1.000 & RoBERTa & 0.961 \\
     M4-Reddit & 0.968 & 0.986 & 0.999 & RoBERTa & 0.907 \\
     M4-wikiHow & 0.767 & 0.899 & 0.998 & RoBERTa & 0.997 \\
     M4-Wiki & 0.964 & 0.983 & 0.996 & RoBERTa & 0.997 \\ \hline
     MULTITuDE & 0.951 & 0.956 & 0.989 & RoBERTa & 0.984 \\ \hline
     MGTBench-N & 0.398 & 0.938 & 1.000 & DistillBERT & 0.948 \\
     MGTBench-S & 0.407 & 0.949 & 1.000 & DistillBERT & 0.989 \\
     MGTBench-T & 0.817 & 0.919 & 1.000 & DistillBERT & 1.000 \\
        \hline
        \end{tabular}\vspace{-3mm}}
\end{table}

\begin{table}[t]
	\caption{TPR and TNR (in \%) in new domains.}\label{tab:otherdataset}
    \setlength{\tabcolsep}{0.5em}
    \vspace{-3mm}
    {
	\begin{tabular}{c|ccc|ccc}
 \hline
      \multirow{2}{*}{} & \multicolumn{3}{c|}{w/o fine-tuning} &\multicolumn{3}{c}{w/ fine-tuning}  \\
		\cline{2-7}
        & Task 1 & Task 2 & Task 3 & Task 1 & Task 2 & Task 3 \\\hline
          \multicolumn{7}{c}{(b) Wiki} \\\hline
         TPR & 100.0 & 99.86 & 98.13 & 99.86 & 98.76 & 94.08 \\
         TNR &  81.13 & 96.50 & 81.13 & 99.23 & 99.54 & 93.85 \\\hline
        \multicolumn{7}{c}{(b) Essay-C} \\\hline
         TPR & 91.09 & 97.13 & 86.82 & 100.0 & 100.0 & 100.0 \\
         TNR & 100.0 & 100.0 & 100.0 & 100.0 & 100.0 & 100.0 \\\hline
        \multicolumn{7}{c}{(c) Essay-P} \\\hline
         TPR & 83.36 & 68.82 & 79.09 & 99.82 & 99.82 & 99.82 \\
         TNR & 99.92 & 99.77 & 99.92 & 99.69 & 98.75 & 99.37 \\\hline
        \multicolumn{7}{c}{(d) BBC News} \\\hline
         TPR & 100.0 & 99.43 & 90.72 & 100.0 & 99.57 & 97.50 \\
         TNR & 99.86 & 99.93 & 99.86 & 100.0 & 99.86 & 98.79 \\\hline
        \end{tabular}\vspace{-3mm}}
\end{table}


\subsection{\checkgpt~Performance Over Time}\label{subsec:transfertovertime}
{
Since its first release, OpenAI has made several major updates to ChatGPT throughout 2023. Hence, we ask the question: \textit{Will \checkgpt~remain effective over time?} We evaluate \checkgpt on ChatLog-HC3 \cite{tu2023chatlog}, which consists of ChaGPT-generated answers for the same questions collected every month starting from 03/2023. In this experiment, we pre-train \checkgpt~with GPABenchmark v1, which was collected before March 2023. As shown in Figure~\ref{fig:chatlog}, \checkgpt's performance has stayed high over the past ten months. Additionally, we have the following observations:

\noindent$\bullet~$ While we only train/tune \checkgpt~with data prior to 03/2023, it yields stable performance over 10 months. Moreover, \checkgpt's accuracy significantly increased for all the domains on 06/2023. A possible explanation is that an update may have introduced more consistent writing styles and stronger patterns to ChatGPT.


\noindent$\bullet~$ \checkgpt~performs worse on OpenQA compared to the original HC3 collected in 12/2022 (Table~\ref{tab:otherdataset}). In a closer look, we find that ChatGPT generated significantly shorter answers in ChatLog. The average length was 133 tokens for HC3-OpenQA but 71 tokens for ChatLog in 03/2023. However, ChatGPT started to generate longer answers later in 2023, with an average length of 98 tokens in December, and the performance of \checkgpt~increases accordingly.

\begin{figure}[!t]
\centering
\includegraphics[width=0.98\columnwidth]{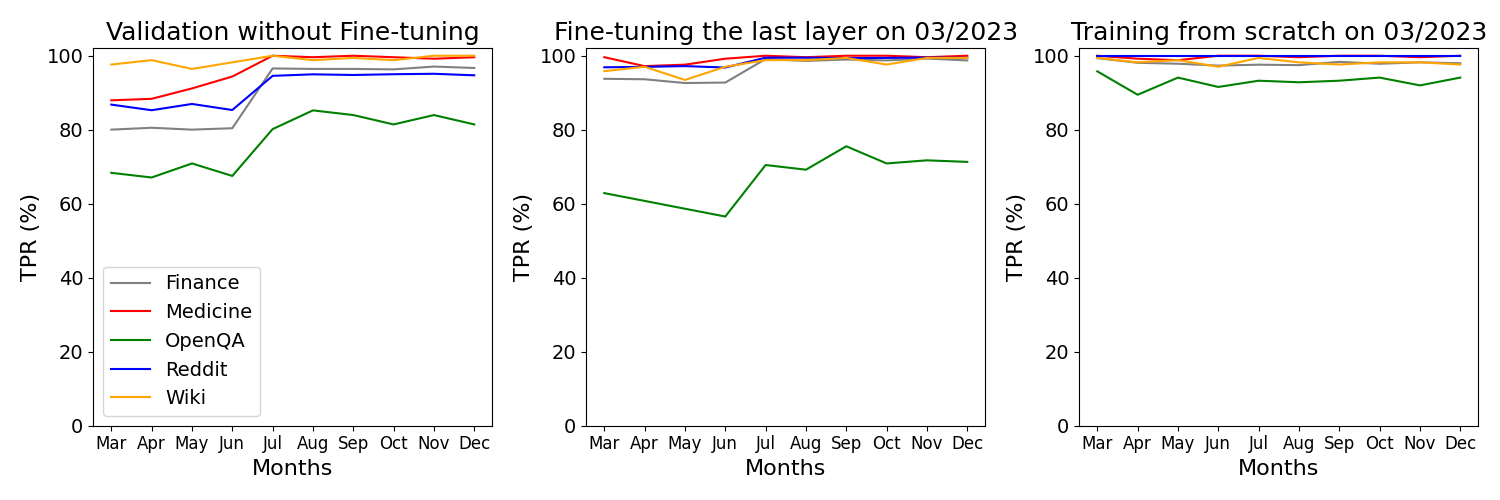}\vspace{-3mm}
\caption{TPR on ChatLog-HC3 with direct validation, fine-tuning, and full re-train.}\vspace{-2mm}
\label{fig:chatlog} \vspace{-2mm}
\end{figure}

}

\subsection{Transferability to New LLMs}\label{subsec:transfertonewmodel}

{
\noindent \textbf{\em GPT-4.} 
We invoke GPT-4, which is the most up-to-date and powerful member of GPT models, with the same prompts in Sec. \ref{sec:subsecgpabenchmark} to generate GPT-WRI, GPT-CPL, and GPT-POL abstracts for  2000 random samples, respectively (small sample size due to strict rate limits). We use the unified classifiers to evaluate all the samples, and \checkgpt~achieves >96\% TPR in all three experiments (99.95\%, 96.90\%, and 96.15\%, respectively).

\noindent \textbf{\em Non-GPT LLMs.} 
Other LLMs might adopt unique model architectures, training datasets, and training methodologies, which may differ significantly from ChatGPT. In this paper, we do not claim or expect the transferability of \checkgpt~to non-GPT LLMs. 
Nevertheless, we still apply \checkgpt~(without tuning) to the content generated by 11 non-GPT LLMs: M4 \cite{wang2023m4}, MULTITuDE \cite{macko2023multitude}, and MGTBench \cite{he2023mgtbench}. As shown in Figure~\ref{tab:otherllms}, \checkgpt~is highly adaptable to content generated by all the LLMs. Note that the F1-scores for models other than ChatGPT in M4 are not reported in \cite{wang2023m4}.


\begin{table}[t]
	\caption{Performance (F1-score) of \checkgpt~on non-GPT LLMs. SOTA: the performance of fine-tuned language models.}\label{tab:otherllms}
    \setlength{\tabcolsep}{0.15em}
    \vspace{-3mm}
    \centering
    {\small 
	\begin{tabular}{c|c|cc|cc|cc}
 \hline
   \multirow{2}{*}{Dataset} & \multirow{2}{*}{Other LLMs} & \multicolumn{2}{c|}{FT-L} & \multicolumn{2}{c|}{FT-A} & \multicolumn{2}{c}{SOTA} \\ \cline{3-6}
   & & Min. & Avg.& Min. & Avg. & Min. & Avg. \\\hline 
     M4-arXiv & \multirow{3}{*}{\parbox{2.5cm}{\centering BloomZ, Cohere, Dolly, Flant-T5/Llama}} & 0.903 & 0.927 & 0.998 & 0.999 & - & -\\
     M4-PeerRead & & 0.820 & 0.868 & 0.996 & 0.999 & - & -\\
     M4-Reddit & & 0.872 & 0.987 & 0.984 & 0.996  & - & -  \\ \cline{2-2}
     M4-wikiHow & \multirow{2}{*}{\parbox{2.5cm}{\centering BloomZ, Cohere, Dolly}} & 0.834 & 0.915 & 0.988 & 0.995   & - & -  \\
     M4-Wiki & & 0.798 & 0.847 & 0.978 & 0.989   & - & - \\ \hline
     \multirow{2}{*}{MULTITuDE} & Alpaca, LLaMA, OPT, 
     & \multirow{2}{*}{0.656} & \multirow{2}{*}{0.758} & \multirow{2}{*}{0.937}& \multirow{2}{*}{0.965}& \multirow{2}{*}{0.883}& \multirow{2}{*}{0.941}
     \\ 
     & OPT-IML, Vicuna & & & & \\\hline
     MGTBench-N & \multirow{3}{*}{\parbox{2.5cm}{\centering ChatGLM, Dolly, StableLM}} & 0.909 & 0.934 & 1.000 & 1.000 &  0.875 & 0.928  \\
     MGTBench-S &  & 0.908 & 0.919 & 1.000 & 1.000 & 0.939 & 0.965 \\
     MGTBench-T &  & 0.920 & 0.956 & 1.000 & 1.000 & 0.966 & 0.988   \\
       \hline 
        \end{tabular}\vspace{-3mm}}
\end{table}

}


\subsection{Advanced Prompt Engineering}\label{sec:prompteng}

As presented in Section \ref{sec:subsecgpabenchmark}, GPABench2 contains 435K additional testing samples using advanced prompt engineering. We evaluate the new dataset with task-specific, discipline-specific, and cross-prompt classifiers. As shown in Table \ref{tab:moreprompt}, \checkgpt's TPRs are consistently high. Moderate decreases are only noticed in LMP, SCP, GKP, MP, and II for Task 2, and LMP for Task 3. However, when a prompt-specific (P1) model is used for the new data, the average TPR decreases by 0.85\%, and the maximum decrease is 8.2\% (detailed in Appendix~\ref{apdx:promptsingle}). This suggests the robustness of the cross-promote models, i.e., the models learned GPT-specific features that are transferable,  instead of prompt-specific bias.

\begin{table}[t]
	\caption{TPR (in \%) for advanced prompts.}\label{tab:moreprompt}
    \setlength{\tabcolsep}{0.39em}
    \centering
    \vspace{-3mm}
    {\small
	\begin{tabular}{c|ccc|ccc|ccc}
 \hline
        & \multicolumn{3}{c}{T1. GPT-WRI} &\multicolumn{3}{c}{T2. GPT-CPL}  &\multicolumn{3}{c}{T3. GPT-POL}\\ \cline{2-10}
         & CS & PHX & HSS & CS & PHX & HSS & CS & PHX & HSS\\\hline
         ZC & 100.0 & 100.0 & 100.0 & 99.38 & 98.91 & 99.21 & 99.79 & 99.84 & 99.79\\
         APE & 100.0 & 100.0 & 99.96 & 99.11 & 99.21 & 99.21 & 99.47 & 99.26 & 99.27\\
         SCP  & 99.95 & 99.94 & 99.98 & 99.15 & 98.43 & 98.67 & 99.64 & 99.81 & 99.73\\
         FSP & 100.0 & 99.98 & 99.92 & 99.68 & 99.61 & 99.44 & 99.45 & 99.24 & 99.54\\
         LMP & 99.94 & 99.98 & 99.94 & 98.60 & 98.97 & 98.85 & 99.01 & 98.99 & 98.89\\
         GKP & 99.96 & 99.98 & 100.0 & 98.50 & 98.62 & 98.70 & 99.78 & 99.73 & 99.80\\
         PP & 100.0 & 100.0 & 99.98 & 99.66 & 99.90 & 99.54 & 99.84 & 99.88 & 99.83\\
         GP  & 99.64 & 99.59 & 99.83 & 99.78 & 99.59 & 99.67 & 99.19 & 99.48 & 99.33\\
         MP & 99.96 & 99.98 & 99.98 & 97.78 & 97.75 & 97.97 & 99.58 & 99.65 & 99.73\\
         II & 100.0 & 99.98 & 99.96 & 99.21 & 98.53 & 99.00 & - & - & -\\ \hline
\end{tabular}}\vspace{0mm}
\end{table}

\subsection{Discussions on \checkgpt's Transferability}\label{dis:insights}\label{sec:discussion}

\begin{figure}[!t]
\centering
\includegraphics[width=\columnwidth]{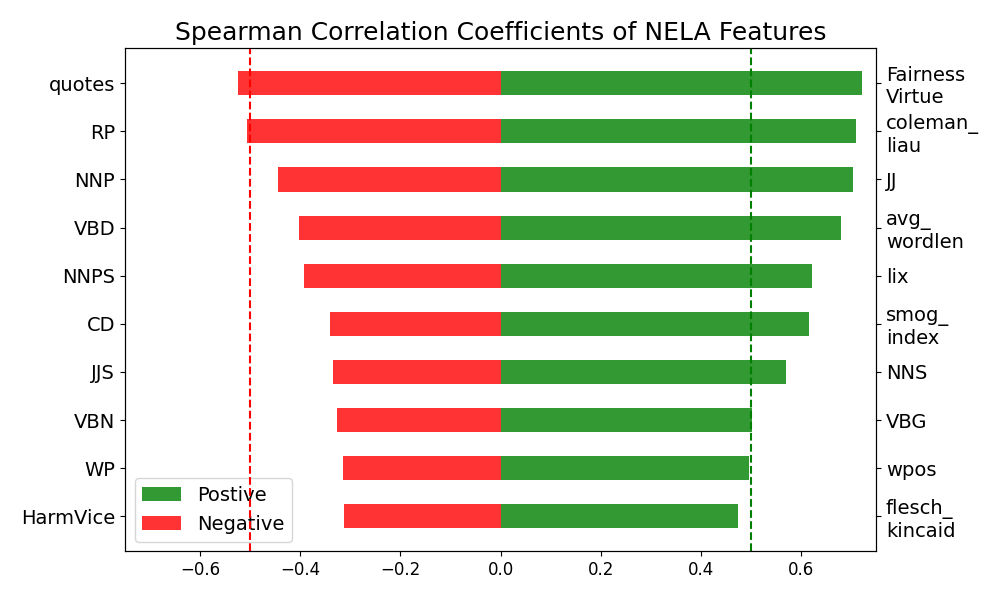} \vspace{-9mm}
\caption{Correlations between domain-specific features and \checkgpt's transferability.} 
\label{fig:spearman}\vspace{-3mm}
\end{figure}

With all the task-, prompt-, model-, and domain-transferability experiments, we raise the critical questions: \textit{What makes \checkgpt~trained on GPABench2 (non-)transferable to other domains}? \textit{How can GPABench2 contribute to general-purpose GPT-text detection}? To investigate this, we again adopt the NELA feature to measure the domain gaps between GPABench2 and all the 39 target domains in the transferability experiments, e.g., Wiki, ASAP Essays, BBC News, etc. We 
examine the correlation of each feature with CheckGPT's transferability, i.e., the performance of \checkgpt~in validations without fine-tuning. We set a threshold at 0.5 on Spearman's Rank Correlation Coefficient. Eventually, we obtain the positive and negative factors influencing the transferability of \checkgpt, as shown in Figure~\ref{fig:spearman}. The top six positive features are (from high to low): Fairness Virtue, Readability (Coleman-Liau Index, average word length, LIX Readability, and Smog Index), adjectives, plural nouns, gerunds. The top two negative factors are quotes and particles. 

Based on the positive factors, we conclude that \textit{\checkgpt~will transfer better to objective, complex, detailed, narrative, or journalistic writing. } Firstly, the fairness virtue indicates an objective tone, while higher readability indicates longer and more complex sentences and word usage. Secondly, descriptive language plays an important role in writing styles with more adjectives. Adjectives add details and expressiveness to the writing by qualifying nouns and pronouns. More adjectives improve the clarity and vividness of the text, indicating creative or detailed writing. Thirdly, writing styles that frequently use plural nouns typically involve discussion of general concepts, discovery, groups, or disciplines rather than focusing on individual or specific instances. They usually serve a technical, scientific, analytical, or political purpose. Lastly, as we have shown in Section~\ref{subsec:baseline}, ChatGPT favors gerunds and \checkgpt~will transfer to descriptive or narrative writing with a sense of continuous actions and processes.

Similarly, based on the negative factors, we conclude that \textit{\checkgpt~will possibly show less transferability to informal writing and conversations, } as particles and quotes will appear more frequently in these writing styles.

\subsection{Use of ChatGPT in arXiv 
Papers} \label{sec:arxivstudy}

\begin{figure}[!t]
\centering
\includegraphics[trim={0 0 0 8mm},clip, width=0.96\columnwidth]{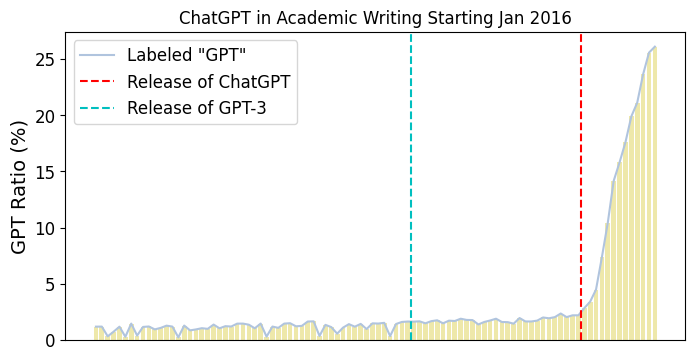}\vspace{-4mm}
\caption{Detecting ChatGPT usage in arXiv papers.}
\label{fig:distlength}\vspace{-4mm}
\end{figure}

Finally, we ask ``\textit{How many authors are using ChatGPT to write/polish their research papers}?'' We collect all the arXiv abstracts in CS from January 2016 to December 2023 ($\sim$450k samples, excluding those in GPABench2). We evaluate each abstract with the unified cross-task cross-prompt cross-disciplinary classifier and show the monthly average positive rates in Figure \ref{fig:distlength}. There is a significant increase in ChatGPT usage with a peak of 26.1\% in December 2023. The average positive rates before, between, and after the releases of GPT-3 and ChatGPT are 1.12\%, 1.78\%, and 7.83\%, respectively. The exponential growth started in December 2022, right after ChatGPT's release on 11/30/2022. Our model also annotates 0.23$\sim$1.66\% of the abstracts posted before GPT3 as GPT-GEN, which may be explained by \checkgpt's 1\% FPR, while LLMs like GPT-2 might also be used by the early adopters.




\subsection{Robustness against Sanitized GPT Output} \label{sec:robustness}
{
We consider the scenario where the user (attacker) sanitizes the raw output from ChatGPT to make it more human-like, in an attempt to escape \checkgpt. We discuss and implement four attacks: 

\noindent\textbf{\em Automatic Rephrasing.} The attacker prompts ChatGPT to rephrase its answers twice. In each prompt, ChatGPT is asked to rephrase its output to make it ``more like human-written.''

\noindent\textbf{\em Mixed Human/GPT Writing (The Mixing Attack).} The attacker substitutes some of the ChatGPT-generated text with human-written text to confuse the detector. In particular, ChatGPT-written abstracts are usually 6-8 sentences long, where the first and last sentences are more generic and less technical. We substitute \textit{each} of the top 3 and last 3 sentences with the corresponding sentence from the HUM abstract. We further attempt to substitute two sentences (top 2 or last 2) from GPT-GEN abstracts. In the end, each attack sample is a hybrid text with GPT-GEN and HUM content.

\noindent\textbf{\em Copyediting.} We make a strong attack assumption that the attacker knows the vocabulary distribution of GPT- and HUM-generated text, and sanitizes GPT text accordingly. To mimic this attack, we examine word and bi-gram distribution in HUM- and GPT-generated abstracts in GBABench2, and identify the words and phrases with the largest frequency discrepancy. i.e., words that are popular in GPT-GEN but rarely used in HUM-WRI. We drop the attributives, and substitute the other words/bigrams that are popular in ChatGPT text with synonyms that are heavily used by humans. Based on the attackers' knowledge and costs, we define three attack levels: Top-3, Top-5, and Top-10. We extract the Top-N words and phrases in each discipline, task, and prompt and combine them in one set. Eventually, there are 76 items in the Top-10 set and 52 items in the Top-3 set. Some examples of substitutions in all three levels include removing or replacing ``this paper'', ``the paper'', ``this study'', ``in this paper'', ``in this study'', ``additionally'', ``furthermore'', etc. Please see Appendix~\ref{apdx:ceaattack} for the detailed vocabulary and substitutions. 

\noindent\textbf{\em Prompt Engineering.} Same as the Copyediting Attack, the attacker knows the term distributions of HUM- and GPT-generated texts. The attacker employs prompt engineering to ask ChaGPT to avoid using the words and phrases that are popular in GPT-GEN text.  

We evaluate our cross-prompt cross-discipline cross-task classifier against 10,000 attack samples for each combination of task, discipline, and prompt. The TPRs are shown in Table~\ref{tab:attacks}. We have the following observations:

\noindent$\bullet~$ \checkgpt's accuracy becomes higher on rephrased text, i.e., nearly 100\% TPR for all the tasks/discipline. It implies that the inherent patterns of ChatGPT get stronger after iterative rewriting.

\noindent$\bullet~$ \checkgpt~is robust when any of the first/last 3 sentences is replaced with a human sentence. Performance for a few tasks dropped to $\sim$75\% when two sentences (about 1/4 to 1/3 of the abstract) are replaced,  while the other tasks still perform >90\%. That is, \checkgpt's detection is not dominated by any single sentence-level indicator. Instead, every sentence contributes to the decision.

\noindent$\bullet~$ \checkgpt's detection accuracy decreases slightly when the unique terms used by ChatGPT (sometimes referred to as ``artifacts'') are removed or replaced with ``human terms''. However, the average TPR remains at 83.1\% (with a minimum of 70.45\%) after removing/replacing 76 top words and phrases in GPT-GEN text. This implies that \checkgpt~learns and utilizes the lexical features in GPT-generated content, but it does not completely rely on them.

\noindent$\bullet~$ \checkgpt~provides the lowest accuracy with Copyediting Attacks in Task 3. In this polishing task, ChatPGT mostly maintains the structure of the paragraph and the sentences but makes word-level tuning. Therefore: (1) The semantic and structural features of GPT-POL abstracts are very similar to HUM abstracts. And (2) the lexicon features of GPT-POL abstracts are effectively eliminated by the manual Copyediting Attack. Both factors contribute to the low accuracy of \checkgpt~against this attack.  


\noindent$\bullet~$ \checkgpt's performance stays high against the Prompt Engineering Attacks. This shows that the ChatGPT's habitual patterns persist even under specially crafted instructions, which aligns with our findings in Section~\ref{sec:prompteng}.

In summary, \checkgpt~is highly robust against \textit{post facto} human interventions of ChatGPT-generated content at word/phrase and sentence levels. As we have observed, \checkgpt~employs many ``weak indicators'' from lexicon, structural, and semantic features that are scattered across the entire text snippet to collectively support a classification decision with relatively strong confidence. 




\begin{table}[t]
	\caption{TPR (in \%) under attacks. CE: Copyediting. PromptEng: Prompt Engineering Attacks. Mixing-F/L: substituting the first/last sentences.}\label{tab:attacks}
    \setlength{\tabcolsep}{0.128em}
    \centering
    \vspace{-1mm}
    {
	\begin{tabular}{c|ccc|ccc|ccc}
 \hline
        & \multicolumn{3}{c}{T1. GPT-WRI} &\multicolumn{3}{c}{T2. GPT-CPL}  &\multicolumn{3}{c}{T3. GPT-POL}\\ \cline{2-10}
         & CS & PHX & HSS & CS & PHX & HSS & CS & PHX & HSS\\\hline
         Rephrasing & 100.0 & 100.0 & 100.0 & 99.99 & 99.98 & 100.0 & 99.99 & 99.97 & 99.95\\ 
         Mixing-F1 & 99.98 & 100.0 & 99.93 & 99.52 & 99.54 & 99.33 & 93.10 & 97.31 & 91.52\\ 
         Mixing-F2 & 99.98 & 100.0 & 99.98 & 99.21 & 99.44 & 99.09 & 95.73 & 98.00 & 95.28\\ 
         Mixing-F3 & 99.89 & 99.97 & 99.77 & 97.59 & 98.62 & 96.83 & 90.52 & 96.22 & 90.73\\ 
         Mixing-L3 & 99.96 & 99.97 & 99.85 & 99.02 & 99.54 & 98.57 & 96.71 & 98.18 & 96.79\\ 
         Mixing-L2 & 99.83 & 100.0 & 99.87 & 98.14 & 99.24 & 98.03 & 94.63 & 98.17 & 94.52\\ 
         Mixing-L1 & 99.80 & 100.0 & 99.78 & 95.21 & 97.33 & 96.24 & 96.44 & 98.39 & 95.83\\ 
          Mixing-F12 & 99.77 & 99.97 & 99.31 & 96.90 & 97.63 & 95.62 & 75.78 & 87.55 & 73.02\\
          Mixing-L12 & 
          95.60 & 98.20 & 96.43
          & 78.05 & 85.57 & 82.67
          & 75.68 & 94.14 & 85.75\\
         CE-Top3 & 96.55 & 98.47 & 92.79 & 93.06 & 95.35 & 91.88 & 83.77 & 91.88 & 85.85\\ 
         CE-Top5  & 92.92 & 96.47 & 88.83 & 89.06 & 92.25 & 87.35 & 78.36  & 89.84 & 81.17\\ 
         CE-Top10 & 89.27  & 92.62  & 82.02 & 85.75 & 88.52  & 82.54 & 70.45 & 84.20 & 72.56\\ 
         PromptEng & 99.99 & 99.94 & 99.99 & 99.19 & 99.02 & 99.62 & 89.92 & 89.30 & 95.92\\ 
         \hline
\end{tabular}}\vspace{-2mm}
\end{table}


%% file: tex/background.tex
\section{Related Work}
\label{sec:background}\label{sec:llm}\label{sec:related}

\noindent\textbf{Neural Language Models and LLM.} Neural networks for word probabilistic modeling have been developed since the 2000s \cite{bengio2000neural, sundermeyer2012lstm, cho2014learning}. Recently, pre-trained language models with general but effective word representation have been widely used, e.g., BERT \cite{devlin2018bert}, RoBERTa \cite{liu2019roberta}, ELMo \cite{emlo2018}, GPT-2 \cite{radford2019language}, and BART \cite{lewis2019bart}. The large language models (LLMs) are trained on massive amounts of data with deep learning frameworks consisting of an ultra-large amount of parameters. ChatGPT is built on top of OpenAI's GPT-3.5 with fine-tuning through supervised and reinforcement learning. 

\begin{table}[t]
    \centering
    \caption{Summary of SOTA LLM-content Detectors.}
    \vspace{-2mm}
    \label{tab:literature}
    \begin{footnotesize}
    \setlength{\tabcolsep}{0em}
    \begin{tabular}{|M{0.35cm}|M{0.6cm}||M{0.5cm}|M{0.45cm}|M{0.55cm}|M{0.55cm}||M{0.8cm}||M{0.7cm}||M{0.55cm}|M{0.45cm}|M{0.6cm}|M{0.5cm}||M{0.7cm}|M{0.6cm}|}
        \hline
        \multicolumn{2}{|c||}{\multirow{2}{*}{Study}} & \multicolumn{4}{c||}{Approach} &  \multirow{2}{*}{\parbox{0.8cm}{Transfer\\-ability}} & \#Hum. & \multicolumn{4}{c||}{Domain}  &  \multicolumn{2}{c|}{Dataset}   \\
        \cline{3-6}\cline{9-14}
         \multicolumn{2}{|c||}{}& Tool & Stat & Hum & DNN &&  Evalu. & News & QA & Essay & Res. & Size & Open\\
        \hline
        
        \hline
        \multirow{5}{*}{\rotatebox[origin=c]{90}{Pre-ChatGPT}}
        & \cite{gehrmann2019gltr} &
          & \ding{108}& &            &
          $-$ & $-$ & 
          \ding{108}& & & \ding{108}           & 
          $300$ &\\ 
         \cline{2-14}
         & \cite{kushnareva2021artificial} &
          & \ding{108}& &  \ding{108}          &
           $-$ & $-$ &
          \ding{108}& & &            & 
          $90k$ & \ding{108}\\ 
          \cline{2-14}
       
        & \cite{mitchell2023detectgpt} &
          \ding{108}& & &            &
           $-$ & $-$ &
          \ding{108} & & \ding{108} &            & 
          $-$ &\\
          \cline{2-14}
           & \cite{theocharopoulos2023detection} &
          & \ding{108}& & \ding{108}           &
           & $-$ &
          & & & \ding{108}           & 
          $28k$ & \ding{108} \\
        \cline{2-14}
         & \cite{zellers2019defending} &
          & & \ding{108} &  \ding{108}          &
           &  $^a$ &
          \ding{108} & & &            & 
          $20k$ & \ding{108} \\ 
         \hline
         
         \hline
         
         \multirow{9}{*}{\rotatebox[origin=c]{90}{ChatGPT}} 
            & \cite{bleuminkkeeping} &
          & & & \ding{108}           &
          & $-$ &
          & & \ding{108} &            & 
          $100k$&\\
         \cline{2-14}
         & \cite{gao2022comparing} &
          \ding{108}& & \ding{108}&            &
          $-$ & $2$ &
          & & &  \ding{108}          & 
          $100$&\\ 
        \cline{2-14}
        & \cite{guo2023close} &
          & \ding{108}& \ding{108}&  \ding{108}          &
          & $17$ &
          & \ding{108} & &            & 
          $125k$&\ding{108}\\
          \cline{2-14}
        
        & \cite{he2023mgtbench} &
          \ding{108}& \ding{108}& & \ding{108}           & \ding{108}
          & &
          & \ding{108} & &            & 
          $6k$&\ding{108}\\
         \cline{2-14}

         & \cite{kumari2023demasq} &
           \ding{108} & \ding{108}& &            &
           &  &
           & \ding{108} &  &     \ding{108}       & 
          $134k$& \\
         \cline{2-14}
         
         & \cite{liu2023argugpt} &
          \ding{108}& \ding{108}& \ding{108}& \ding{108}           &
          & $43$ &
          & & \ding{108} &            & 
          $8k$&\ding{108}\\
         \cline{2-14}

         & \cite{macko2023multitude} &
          \ding{108}& \ding{108}& & \ding{108}           &
          & &
         \ding{108} & &  &            & 
          $74k$&\ding{108}\\
         \cline{2-14}

         & \cite{wang2023m4} &
          & \ding{108}& & \ding{108}           &
          \ding{108} &  &
          \ding{108} & \ding{108} & \ding{108} &     \ding{108}       & 
          $122k$&\ding{108}\\
         \cline{2-14}

         & Ours &
          \ding{108}& \ding{108}& \ding{108}& \ding{108}           &
          \ding{108} & {\color{blue}{\textbf{242$^b$}}} &
          \ding{108} & & \ding{108} &  \ding{108}          & 
          {\color{blue}{\textbf{>2.8m}}}&\ding{108}\\
         \cline{2-14}
         \hline

         \hline
    \end{tabular}\\
\begin{flushleft}
Pre-ChatGPT: Grover and GPT-2/3. Tool: used/evaluated online detection tools. Sat: Statistical approach. Hum: human evaluation. \# Hum. Evalu.: number of human evaluators. Res.: research papers/abstracts. Open: open-sourced.
\\$^a$ The number of human evaluators is not explicitly provided.
\\$^b$ 155 evaluators in the first experiment and 87 evaluators in the second experiment. 
\end{flushleft}
\vspace{-4mm}
\end{footnotesize}
\end{table}

\noindent\textbf{LLM-Content Detection.}  The detection of LLM-texts can be categorized into white-box and black-box approaches \cite{tang2023science}. 
Prior works \cite{gehrmann2019gltr, kushnareva2021artificial, mitchell2023detectgpt, theocharopoulos2023detection, zellers2019defending} study LLM-content detection for pre-ChatGPT models. \cite{gao2022comparing} evaluated 50 ChatGPT-written biomedical research abstracts with human reviewers and a RoBERTa-based classifier. Their findings show that 34\% of the abstracts are labeled as likely human-written. \cite{bleuminkkeeping} trained a transformer-based deep learning model to distinguish between AI-generated and human-written essays in a range of different education levels. \cite{guo2023close} and \cite{liu2023argugpt} conduct comprehensive studies, including human evaluators. \cite{guo2023close} analyzes a series of question-answering datasets in both English and Chinese, and \cite{liu2023argugpt} targets essays written by students and English learners. \cite{he2023mgtbench} establishes the first machine-generated text benchmark evaluating a number of detection approaches. 

The detailed comparisons of existing GPT detectors are listed in Table~\ref{tab:literature}. Compared with the other approaches, \checkgpt~(1) collects/uses a significantly larger dataset, (2) uses a model-agnostic design for higher affordability, upgradability, and flexibility, and (3) achieves very high accuracy, transferability, and interoperability. 

DEMASQ \cite{kumari2023demasq} is the most recent work that is similar to \checkgpt. It designed an energy-based model to identify the inherent differences between GPT- and human-generated text. Experiments across 8 sub-datasets achieved an accuracy of 73.9\% to 100\%. DEMASQ employs a very different attack model as it assumes alterations to ChatGPT output made by humans
to evade detection. Its detailed alteration algorithm, datasets, and models are not yet open-sourced, therefore, we are unable to run a comparative study. 


\noindent\textbf{Security and Ethics in AIGC Application.} AI-generated content (AIGC) has been used in adversarial activities before LLMs were introduced \cite{ferrara2016rise}, while ChatGPT may provide a powerful tool to malicious actors \cite{renaud2023chatgpt,derner2023beyond}. The detection of AI-backed bots, scams, and misinformation has been extensively studied in the literature  \cite{cresci2020decade,shahid2022detecting,yang2020scalable,lu2022effects}, while the rise of ChatGPT introduces both new opportunities \cite{heidari2021bert,heidari2020using,dukic2020you,garcia2019empirical,hoes2023using,wang2023bot} and challenges \cite{gradonm2023electric,de2023chatgpt,mansfield2023weaponising,grbic2023social,roy2023generating,hazell2023large,cambiaso2023scamming}. While Open AI has enforced internal mechanisms to prohibit the unethical use of ChatGPT, the restrictions could be evaded \cite{li2023multi,liu2023jailbreaking}. Finally, there are also discussions and concerns with ChatGPT's potential impact on education and research \cite{stokel2022ai,willems2023chatgpt, firat2023chatgpt}, especially on authorship and plagiarism  \cite{stokelchatgpt,flanagin2023nonhuman,khalil2023will,anders2023using}.  

%% file: tex/conclusion.tex
\section{Conclusion}\label{sec:conclusion}

In this paper, we first present GPABench2, a benchmarking dataset with 2.385 million samples of human-written, GPT-written, GPT-completed, and GPT-polished research paper abstracts. Next, we show that the existing ChatGPT detectors and human users are incapable of identifying GPT-content in GPABench2. We further present \checkgpt, a deep learning-based detector for GPT-generated academic writing. With extensive experiments, we show that \checkgpt~is highly accurate, affordable, flexible, and transferable.

%% file: tex/appendix.tex
\appendix

\section{Prompts used in GPABench2}
\subsection{Prompts Used in the Main Dataset of GPABench2}\label{apdx:prompt3}
The complete prompts used in GPABench2 data collection are listed as follows:

\begin{enumerate}[noitemsep, topsep=1mm, left=0pt]
\item Prompt 1: zero-shot prompt.

\noindent\textbf{Task 1}: Here is the title of an academic research paper. Please write a paper abstract about it: \{input\}.

\noindent\textbf{Task 2}: Here is the first half of the abstract of an academic research paper. Please complete its second half with approximate \{X\} words: \{input\}.

\noindent\textbf{Task 3}: Here is the abstract of an academic research paper. Please rewrite it for clarity: \{input\}.

\item Prompt 2: Prompt with context. 

\noindent\textbf{Task 1}: Write an abstract of a research paper in \{discipline\} with first-person, clear, and academic language about "\{title\}".

\noindent\textbf{Task 2}: Write a well-written and coherent continuation, with approximately \{X\} words, of the following first half of the abstract of a research paper in \{discipline\}: "{input}"

\noindent\textbf{Task 3}: Write a polished and refined version of the following abstract of a research paper in \{discipline\} to improve its overall quality and readability: "\{input\}"

\item Prompt 3: Role-playing prompt. 

\noindent\textbf{Task 1}: I want you to act as an academic paper writer. You are familiar with the topics in \{discipline\}. You will be responsible for writing a paper abstract. Your task is to generate an abstract for a paper with a given title. Please only include the written abstract in your answer. Here is the title of the paper: "\{input\}"

\noindent\textbf{Task 2}: I want you to act as an academic paper writer. You are familiar with the topics in \{discipline\}. You will be responsible for completing an unfinished paper abstract. Your task is to create a seamless and well-written continuation with approximately \{X\} words for the second half, given the provided first half of the abstract. Please only include the second half in your answer. Here is the first half of the abstract: "\{input\}"

\noindent\textbf{Task 3}: I want you to act as an academic paper writer. You are familiar with the topics in \{discipline\}. You will be responsible for rewriting a paper abstract. Your task is to improve the writing and clarity of the abstract. Please only include the rewritten abstract in your answer. Here is the original abstract of the paper: "\{input\}"

\item Prompt 4: detailed user requirements and instructions. 

\noindent\textbf{Task 1}: Please act as an expert paper writer and write the abstract section of a paper from the perspective of a paper reviewer to make it fluent and elegant. Please only include the written abstract in your answer. Here are the specific requirements: 1. Enable readers to grasp the main points or essence of the paper quickly. 2. Allow readers to understand the important information, analysis, and arguments throughout the entire paper. 3. Help readers remember the key points of the paper. 4. Please clearly state the innovative aspects of your research in the abstract, emphasizing your contributions. 5. Use concise and clear language to describe your findings and results, making it easier for reviewers to understand the paper. Here is the title of the paper: "\{input\}"

\noindent\textbf{Task 2}: Please act as an expert paper writer and complete the second half of the given first half of an abstract section from the perspective of a paper reviewer to make it fluent and elegant. Please only include the second half of the abstract in your answer. Here are the specific requirements: 1. The length of the second half should be about \{X\} words. 2. The existing content should serve as the foundation, and the new portion should seamlessly integrate with it. 3. Use your expertise and maintain its technical accuracy and clarity. 4. Ensure a coherent and logical flow between the first and second halves. 5. Use clear and academic language, making it easier for reviewers to understand the paper. Here is the first half of the paper abstract section: "\{input\}"

\noindent\textbf{Task 3}: Please act as an expert paper editor and revise the abstract section of the paper from the perspective of a paper reviewer to make it more fluent and elegant. Please only include the revised abstract in your answer. Here are the specific requirements: 1. Enable readers to grasp the main points or essence of the paper quickly. 2. Allow readers to understand the important information, analysis, and arguments throughout the entire paper. 3. Help readers remember the key points of the paper. 4. Please clearly state the innovative aspects of your research in the abstract, emphasizing your contributions. 5. Use concise and clear language to describe your findings and results, making it easier for reviewers to understand the paper. Here is the original abstract section of the paper: "\{input\}"

\end{enumerate}

\subsection{Prompts used in the Additional Testing Samples}\label{apdx:otherprompt}
The details of the prompt techniques covered in Sec~\ref{sec:discussion} are as follows.

\begin{enumerate}
    \item \textbf{Zero-shot Chain-of-Thought Prompting (ZC).} Zero-shot Chain-of-Thought (Zero-shot CoT) Prompting \cite{zeroshot} utilizes a trigger phrase like "Let's think step by step." to guide the model through a sequence of necessary reasoning steps for the problems. Each prompt has two parts: the first generates a chain of thought, and the second extracts the final answer. In our experiment, we adhered the trigger phrase to our original prompts.
    
    \item \textbf{Automatic Prompt Engineer (APE).}  APE \cite{ape} automates the process of generation and selection of the prompts for LLMs with an iterative scoring and resampling mechanism. We simplify this process by directly adopting the optimal trigger phrase "Let's work this out in a step by step way to make sure that we have the correct (good) answer." from \cite{ape}. 
    
    \item \textbf{Self-critique Prompting (SCP).} This method \cite{critique} engages LLMs in a self-evaluation process to enhance model performance \cite{critique1, critique2,critique3,critique4}. The LLMs provide self-reflective feedback or suggestions on their own responses and improve them. In our experiment, we instruct ChatGPT to perform self-critique and self-improvement subsequently.
    
    \item \textbf{Few-shot Prompting (FSP).} Few-shot prompting \cite{fewshot}, also widely recognized as few-shot in-context learning, involves a set of demonstrations or examples to condition the LLMs to the context. In our experiment, we provide three paper abstracts each time to facilitate ChatGPT's understanding of academic writing styles.
    
    \item \textbf{Least-to-Most Prompting  (LMP).} This method \cite{least} consists of two stages: decomposing the problem into easier subproblems and solving them subsequently. In our experiment, we asked ChatGPT to decompose our original question and respond following the devised recipe. 
    
    \item \textbf{Generated Knowledge Prompting (GKP).}  Generated Knowledge Prompting \cite{generated} includes two stages: initial queries asking the LLM to give relevant information, which is subsequently refined as the context for further instructions. This recursive prompting technique leverages the LLM’s knowledge-generation capability.
    
    \item \textbf{GPT-generated Prompts (GP).} Following \cite{Creator}, we appoint ChatGPT as a prompt generator. We assign the task of drafting and improving the prompts to optimally align with user needs while ensuring their clarity, conciseness, and comprehensibility for ChatGPT. Here is the prompt used in this paper:
    "I want you to become my prompt generator. Your goal is to help me craft the best possible prompt for my needs. The prompt will be used by you, ChatGPT. You will follow the following process: 1. Your first response will be to ask me what the prompt should be about. I will provide my answer, but we will need to improve it. 2. Based on my input, you will generate the revised prompt. It should be clear, concise, and easily understood by you."
    
    \item \textbf{Prompt Perfect (PP).} Prompt Perfect \cite{promptperfect}, a third-party plugin supported in the OpenAI GPT-4 interface \cite{chatgptplugin}, rephrases user inputs to improve the quality of ChatGPT's responses. In our experiments, we use Prompt Perfect to rephrase our original prompt.

    \item \textbf{Meta Prompts (MP).} Similar to self-critique prompting, meta prompts instruct LLMs to revise both the answer and the prompt \cite{metaprompt}. At the end of the process, LLMs generate an additional response based on the refined prompt.

    \item \textbf{Instruction Induction (II).} This method \cite{honovich2022instruction} searches the natural language space for an apt description of the target task. It introduces a paradigm where the model is provided with a few input-output pairs and then prompted to infer a fitting instruction. Task 3 was omitted in our experiments due to the lack of abstracts before and after proficient polishing. For Task 1 and 2, we use the original title and abstract as examples. The prompts inferred by ChatGPT are "Given the title of a research paper, please generate an abstract that outlines the main contributions, methodology, and results of the paper." and "Given an abstract or introduction discussing the motivation and problem definition of a research paper, provide a continuation which describes the proposed solution, methodology, and results.".
                       
\end{enumerate}

\section{Benchmarking Open-source and Online ChatGPT Detectors}\label{apdx:benchmarking}

\subsection{Open-source Detectors used in This Study}\label{appdx:detectordetails}

The following open-source detectors for LLM and CharGPT are adopted in the benchmarking study in this paper (Section \ref{sec:benchmark}). 

\noindent
\textbf{HC3-Perplexity (HC3-PPL).} This is the pre-trained perplexity-based model on the HC3 dataset \cite{guo2023close}.

\noindent
\textbf{HC3-GLTR.} This is the pre-trained HLR (GLTR Test 2) model on the HC3 dataset \cite{guo2023close}.

\noindent
\textbf{HC3-Roberta (HC3-RBT).} This is the RoBERTa parameters fine-tuned on HC3 dataset \cite{guo2023close}.

\noindent
\textbf{OpenAI-Roberta (OpenAI-RBT).} This is RoBERTa parameters fine-tuned on GPT-2 model outputs by OpenAI \cite{solaiman2019release, openairoberta}.

\noindent
\textbf{Rank.} This method computes the average rank of each observed token by likelihood given a language model. Text with a smaller average rank is likely machine-generated \cite{mitchell2023detectgpt}.

\noindent
\textbf{Log-Rank.} The average log-ranks of each token are computed applying logarithm \cite{mitchell2023detectgpt}, which outperforms rank-based detection.

\noindent
\textbf{Histogram-of-Likelihood Ranks (HLR).} GLTR \cite{gehrmann2019gltr} uses this as Test 2. With the ranks of each token, HLR further binned them into 4-dimensional histograms using three thresholds.

\noindent
\textbf{Total Probability (TP).} \cite{solaiman2019release} observes that the likelihood of machine-generated text sequences differs from human-written text according to a language model. The average token-wise log probability of the machine-generated text is usually higher.

\noindent
\textbf{Perplexity (PPL).} Perplexity is the exponentiated average negative log-likelihood of a text input given a Language Model. \cite{guo2023close} observes that the perplexity is lower on GPT-generaated content.

\noindent
\textbf{Entropy.} The entropy-based detection \cite{gehrmann2019gltr, mitchell2023detectgpt} follows the hypothesis that the machine-generated text will be more "in-distribution" \cite{gehrmann2019gltr} or "out-of distribution" \cite{mitchell2023detectgpt}, resulting in a difference of confidence given a pre-trained language model.

\noindent
\textbf{DetectGPT.} \cite{mitchell2023detectgpt} proposes a zero-shot method based on the observation that text generated by LLMs typically resides in regions of negative curvature, considering the log probability. Thus, it perturbs the input text with small changes and analyzes the change in likelihood. A significant drop indicates machine-generated text.



\subsection{Benchmarking GPTZero}
\noindent\textbf{GPTZero.} For each text paragraph, GPTZero \cite{gptzero} reports a binary decision of ``human'' or ``GPT''. As shown in Table \ref{tab:otherdetectors} (a), GPTZero demonstrates very high accuracy with human-written abstracts (98.1\% average accuracy across all the topics). However, its detection accuracy for GPT-generated abstracts appears to be very low, with an average accuracy of 24.3\%. That is, GPTZero has a very strong tendency to classify an input abstract as ``human-written''. From Task 1 to Task 3, the detection performance decreases significantly (from 42.5\% to 8.1\%). That is, when more information is given to ChatGPT, the generated text appears to be more ``human-like'' in the eyes of GPTZero.

\subsection{Benchmarking ZeroGPT}\label{apdx:zerogpt}


For each input text snippet, ZeroGPT reports a decision from nine different labels. We map them to integer scores in the range of [0, 8], where 0 indicates human-written and 8 indicates AI/GPT-generated. 

For each input text snippet, ZeroGPT \cite{zerogpt} returns one of the nine possible decisions. We assign an integer score of [0, 8] as follows:

\begin{enumerate}[noitemsep, topsep=1mm, left=0pt]
\setcounter{enumi}{-1}
    \item Your text is Human written
    \item Your text is Most Likely Human written
    \item Your text is Most Likely Human written, may include parts generated by AI/GPT
    \item Your text is Likely Human written, may include parts generated by AI/GPT
    \item Your text contains mixed signals, with some parts generated by AI/GPT
    \item Your text is Likely generated by AI/GPT
    \item Your text is Most Likely AI/GPT generated
    \item Most of Your text is AI/GPT Generated
    \item Your text is AI/GPT Generated
\end{enumerate}

The distribution of the scores for each task and each discipline is shown in Table \ref{tab:zerogpt}. For instance, for GPT-polished abstracts (Task 3) in CS, 88.3\% were annotated as ``human'' by ZeroGPT, while 4.7\% were annotated as ``Most likely human written''. 

When we converted the 9-point scores to binary decisions of ``GPT''/``Human'', a threshold of 4 was used. While we can also make the case that categories 2, 3, 4 should be categorized as ``GPT'' in Task 3, since the decision statements indicate that they ``may include parts generated by AI/GPT,'' which is the case for Task 3. However, changing the decision threshold will not significantly change the observations and conclusions in Section \ref{sec:comparison}, since only a very small portion of the samples in Task 3 were annotated with those three labels, as shown in Table \ref{tab:zerogpt}. For Tasks 1 and 2, the text snippets we sent to ZeroGPT were completely written by ChatGPT, hence, a threshold of 4 is the most reasonable choice. 

ZeroGPT's average detection accuracy for each task and each discipline was presented in Table \ref{tab:otherdetectors} (b1), and the average score for each experiment in Table \ref{tab:otherdetectors} (b2). ZeroGPT's detection accuracy for human-written abstracts is close to 100\% in CS and physics, and slightly lower ($\sim$95\%) in humanities and social sciences (HSS). Its accuracy with fully GPT-written abstracts is also high, especially for HSS (92.3\%). However, the detection accuracy for GPT-completed and GPT-polished  abstracts in CS and physics appears to be very low (in the range of [5\%, 25.3\%]), while the accuracy for HSS appears to be relatively higher. While ZeroGPT claims a detection accuracy of 98\%, it appears to be less effective in academic writing. Similar to GPTZero, ZeroGPT also has a tendency to classify GPT-generated text as human-written.

\begin{table}[t]
	\setlength{\tabcolsep}{0.25em}
 \caption{Distribution of detection score generated by the ZeroGPT: 0: human-written; 8: GPT-generated. The largest score category for each experiment is shown in bold. }\label{tab:zerogpt} 
 {\small
	\begin{tabular}{c|ccc|ccc|ccc}
        \hline
      \multirow{2}{*}{} & \multicolumn{3}{c|}{T1. GPT-ERI} &\multicolumn{3}{c|}{T2. GPT-CPL}  &\multicolumn{3}{c}{T3. GPT-POL}\\
		\cline{2-10}
        & CS & PHX & HSS & CS & PHX & HSS & CS & PHX & HSS\\\hline\hline
    \multicolumn{10}{c}{(a) Score distribution (in \%) of GPT-generated abstracts.} \\\hline
0	&	16.7	&	21	&	1.7	&	\textbf{52.7}	&	\textbf{75.7}	&	18	&	\textbf{88.3}	&	\textbf{93}	&	\textbf{52}\\
1	&	4.7	&	3	&	2	&	1.3	&	1	&	1	&	4.7	&	2	&	6.7	\\
2	&	6	&	5	&	2	&	13.3	&	11.3	&	13.7	&	2	&	1	&	8.3	\\
3	&	2.7	&	1	&	2	&	6.7	&	0.7	&	3	&	1.3	&	0.7	&	5.3	\\
4	&	2.7	&	1.7	&	0	&	0.7	&	1.3	&	2	&	0.3	&	0.7	&	3	\\
5	&	4.3	&	0.7	&	0.3	&	5.3	&	2	&	7.7	&	1.3	&	0	&	6.7	\\
6	&	4.7	&	5.7	&	2.7	&	4	&	3.3	&	7.3	&	1	&	0.7	&	5	\\
7	&	8.7	&	17.3	&	3.3	&	3.3	&	1	&	9.7	&	0	&	0.3	&	3.3	\\
8	&	\textbf{49.7}	&	\textbf{44.7}	&	\textbf{86}	&	12.7	&	3.7	&	\textbf{37.7}	&	1	&	1.7	&	9.7	\\ \hline\hline
        \multicolumn{10}{c}{(b) Score distribution (in \%) of human-written abstracts.} \\\hline
0	&	\textbf{93.7}	&	\textbf{97.7}	&	\textbf{79}	&	\textbf{96.3}	&	\textbf{98.7}	&	\textbf{87.3}	&	\textbf{92}	&	\textbf{96}	&	\textbf{79}	\\
1	&	3.3	&	0	&	9	&	0.7	&	0	&	0.7	&	4	&	1	&	6.7	\\
2	&	3	&	0.7	&	5	&	2.7	&	1	&	5.7	&	2	&	1.3	&	5	\\
3	&	0	&	0	&	2	&	0	&	0	&	1	&	0.3	&	0.3	&	2	\\
4	&	0	&	0.3	&	1.7	&	0	&	0	&	1.3	&	0.3	&	0	&	1.7	\\
5	&	0	&	0	&	1	&	0	&	0.3	&	1	&	0.3	&	0.3	&	2	\\
6	&	0	&	0	&	1.3	&	0	&	0	&	1	&	0	&	0	&	2	\\
7	&	0	&	0.3	&	0.7	&	0.3	&	0	&	0.3	&	0	&	0.3	&	0.3	\\
8	&	0	&	1	&	0.3	&	0	&	0	&	1.7	&	1	&	0.7	&	1.3	\\
\hline\hline
        \end{tabular}}

\end{table}

\subsection{Benchmarking OpenAI's Text Classifier}\label{apdx:openai}

For each input text snippet, the OpenAI text classifier \cite{openaidetector} returns a decision from one of the five classes. We map them to an integer score of [0, 4] as follows:

\begin{enumerate}[noitemsep, topsep=1mm, left=0pt]
\setcounter{enumi}{-1}
    \item The classifier considers the text to be very unlikely AI-generated.
    \item The classifier considers the text to be unlikely AI-generated.
    \item The classifier considers the text to be unclear if it is AI-generated.
    \item The classifier considers the text to be possibly AI-generated.
    \item The classifier considers the text to be likely AI-generated.
\end{enumerate}

The distribution of the scores for each task and each discipline is shown in Table \ref{tab:openaidetector}. For instance, for human-written CS abstracts, 11\% are classified as ``very unlikely AI-generated, 40\% are classified as ``unlikely AI-generated'', 45.3\% are classified as ``unclear if it is AI-generated'', and the remaining 3.7\% are classified as ``possibly AI-generated''.

We use a threshold of 2 to generate a binary decision for each test. Note that a classification of ``(2) unclear if it is AI-generated'' is considered wrong for both GPT-generated and human-written inputs. We present OpenAI's classification accuracy in Table \ref{tab:otherdetectors} (c1) and the average scores in Table \ref{tab:otherdetectors} (c2). OpenAI's classifier shows slightly different patterns from GPTZero and ZeroGPT. It demonstrates moderate performance in classifying abstracts that are fully written by humans or GPT. However, its  accuracy for GPT-completed and GPT-polished abstracts appears inadequate (but slightly better than GPTZero and ZeroGPT). We also noticed that this classifier is very sensitive to the length of text. While it requires a minimum of 1,000 characters for each input text snippet, a shorter input (e.g., input in Task 2 GPT-CPL) is more likely to yield a wrong or ``unclear'' decision.

Note that OpenAI has taken its detector offline in July 2023, ``\textit{due to its low rate of accuracy}.'' \footnote{\url{https://openai.com/blog/new-ai-classifier-for-indicating-ai-written-text}} This is another indication that distinguishing human-written and GPT-generated text is a very challenging task even for the owner of GPT.   

\begin{table}[!t]
	\setlength{\tabcolsep}{0.25em}
 \caption{Distribution of detection score generated by the OpenAI text classifier: 0: very unlikely AI-generated; 2: unclear if it is AI-generated; 4: likely AI-generated. The largest score category for each experiment is shown in bold. }\label{tab:openaidetector} 
 {
	\begin{tabular}{c|ccc|ccc|ccc}
        \hline
      \multirow{2}{*}{} & \multicolumn{3}{c|}{T1. GPT-ERI} &\multicolumn{3}{c|}{T2. GPT-CPL}  &\multicolumn{3}{c}{T3. GPT-POL}\\
		\cline{2-10}
        & CS & PHX & HSS & CS & PHX & HSS & CS & PHX & HSS\\\hline\hline
    \multicolumn{10}{c}{(a) Score distribution (in \%) of GPT-generated abstracts.} \\\hline
0	 & 	0	 & 	0	 & 	0	 & 	0	 & 	0	 & 	8	 & 	4	 & 	5	 & 	11.3	\\
1	 & 	0.3	 & 	0.3	 & 	3.3	 & 	1.3	 & 	12.3	 & 	11	 & 	23.7	 & 	35.3	 & 	31.3	\\
2	 & 	19	 & 	29.7	 & 	33.7	 & 	35	 & 	\textbf{64}	 & 	\textbf{53.7}	 & 	\textbf{66}	 & 	\textbf{55.3}	 & 	\textbf{51.3}	\\
3	 & 	\textbf{50}	 & 	\textbf{50.7}	 & 	\textbf{51}	 & 	\textbf{56}	 & 	22.7	 & 	24	 & 	6.3	 & 	4	 & 	6	\\
4	 & 	30.7	 & 	19.3	 & 	12	 & 	7.7	 & 	1	 & 	3.3	 & 	0	 & 	0.3	 & 	0 \\ \hline\hline
        \multicolumn{10}{c}{(b) Score distribution (in \%) of human-written abstracts.} \\\hline
0	 & 	11	 & 	15.7	 & 	\textbf{60}	 & 	4.3	 & 	7.7	 & 	\textbf{56.3}	 & 	12.7	 & 	18	 & 	\textbf{62.0}	\\
1	 & 	40	 & 	\textbf{54}	 & 	24	 & 	31	 & 	\textbf{52}	 & 	23.3	 & 	38	 & 	\textbf{51}	 & 	26	\\
2	 & 	\textbf{45.3}	 & 	28.3	 & 	14	 & 	\textbf{54}	 & 	38	 & 	16.7	 & 	\textbf{48.3}	 & 	29.7	 & 	10.7	\\
3	 & 	3.7	 & 	2	 & 	1.3	 & 	10.7	 & 	2	 & 	3.3	 & 	1	 & 	1.3	 & 	1	\\
4	 & 	0	 & 	0	 & 	0.7	 & 	0	 & 	0.3	 & 	0.3	 & 	0	 & 	0	 & 	0.3	\\
\hline\hline

        \end{tabular}}

\end{table}

\section{Other Datasets}\label{apdx:newdomain}
Note that the same data samples are used for testing before and after fine-tuning in Section~\ref{subsec:newdomain}.
\begin{itemize}[noitemsep, topsep=0pt, left=0pt]
    \item \textbf{Wikipedia Abstracts}. The dataset contains the first introductory section of Wiki articles. We revise the ChatGPT prompts to avoid terms such as ``research'' and  ``paper''. For example, we use the prompt ``\textit{Please generate a brief introduction of ...}'' in Task 1.
    \item \textbf{ASAP Essays}. We use two types of essays from the Hewlett Foundation Automated Essay Scoring dataset \cite{hewlettessaydataset}: [Essay-C] Essay set 1 contains 1,785 essays of 350 words on average. We adopt the original prompt from the dataset in Task 1: ``\textit{Write a letter to your local newspaper in which you state your opinion on the effects computers have on people. Persuade the readers to agree with you.}''. [Essay-P] Essay set 7 contains 1,730 stories about patience. We refer to the original prompts from the dataset to design ChatGPT prompts e.g., ``\textit{write a story in your own way about patience}'' is used in Task 1. We design prompts for Tasks 2 and 3 accordingly. We remove essays that are shorter than 70 words. 
    \item \textbf{BBC News Article Dataset}. This dataset contains 1,454 BBC news articles from 2004 to 2005 in five topical areas: business, entertainment, politics, sport, and technology \cite{greene06icml}. We use prompts to emphasize ``news articles'' to ChatGPT, e.g., ``\textit{Please generate a news article titled ...}''. 
\end{itemize}

\begin{table}[!t]
\caption{TPR (in \%) of prompt-specific models on advanced prompts.}\label{tab:morepromptsingle}
    \setlength{\tabcolsep}{0.25em}
    \centering
    \vspace{-2mm}
    {
\begin{tabular}{c|ccc|ccc|ccc}
\hline
        & \multicolumn{3}{c}{T1. GPT-WRI} &\multicolumn{3}{c}{T2. GPT-CPL}  &\multicolumn{3}{c}{T3. GPT-POL}\\\hline
         & CS & PHX & HSS & CS & PHX & HSS & CS & PHX & HSS\\\hline
ZC & 99.96 & 100.00 &  99.98 &  98.67 &  97.81 &  98.60 &  98.01 & 99.53 &   99.08 \\
APE & 99.96 &  99.98 &  99.90 &  98.14 & 98.06 & 98.69 &  97.05 &  97.81 & 98.00 \\
SCP & 99.91 & 99.75 & 99.81 & 97.78 & 97.40 & 97.56 &  98.35 &  99.28 & 99.08 \\
FSP & 99.88 & 99.98 & 99.87 & 99.01 & 98.82 & 99.02 &  97.19 &  97.93 & 98.50 \\
LMP & 99.82 & 99.88 & 99.64 & 96.95 & 97.19 & 97.93 &  95.47 &  97.13 & 97.49 \\
GKP & 99.27 & 99.93 & 99.98 & 96.74 & 97.26 & 97.87 &  98.08 &  99.17 & 99.37 \\
PP & 94.70 & 100.00 & 99.96 & 98.71 & 99.40 & 99.00 &  98.99 &  99.48 & 99.31 \\
GP & 98.75 & 99.03 & 98.71 & 99.23 & 99.24 & 99.48 & 97.61 &  99.03 & 98.68 \\
MP & 99.78 & 99.90 & 99.83 & 95.43 & 96.38 & 97.01 & 97.92 &  99.46 & 99.12 \\
II & 99.95 & 99.88 & 99.90 & 96.20 & 96.18 & 90.79 & - & - & - \\
\hline
\end{tabular}}\vspace{-3mm}
\end{table}

\section{Additional Experimental Results}


\subsection{NELA Feature Importance} \label{apdx:nelagb}
{

Figure~\ref{fig:nelatask1}, \ref{fig:nelatask2}, and \ref{fig:nelatask3} shows the NELA feature importance for each task over disciplines.
For Task 1, readability (Coleman-Liau Index \cite{colemanindex}) is always the dominating feature, regardless of the discipline. For example, the average Coleman-Liau index for computer science is 20.63 for GPT-WRI abstracts and 15.44 for HUM abstracts. This shows that ChatGPT tends to compose longer sentences and use longer words during writing, which makes the text harder to comprehend than human writing. 
For Task 2, Coleman-Liau's readability is still the most important feature for all three disciplines, and its importance is far larger than the others. Other contributing factors include the gerunds or present verbs (for all three disciplines), Lix readability (for CS and PHX), possessive pronouns (for PHX), and Flesch–Kincaid readability (for HSS). For PHX, the average number of possessive pronouns is 0.05 for HUM abstracts, 0.18 for GPT-WRI abstracts, and 0.15 for GPT-CPL abstracts. Considering gerund or present verbs, the average numbers are 0.025, 0.017, and 0.024 for HUM abstracts in CS, PHX, and HSS, and these numbers nearly double for GPT-CPL abstracts. A possible explanation is that ChatGPT habitually uses more gerunds as adverbials. 
For Task 3, the importance is more scattered among features. Readability and determiners are important across three disciplines. Gerunds and personal pronouns are important for CS and PHX. Specifically, ChatGPT uses more determiners, more gerunds, and fewer personal pronouns than human writers after polishing. Additionally, ChatGPT uses fewer non-third-person singular present verbs for PHX and HSS and more third-person singular present verbs for CS. 
However, readability is consistently the most important feature, and most other differences (except gerunds) are subtle. Some of the patterns are unique and not general across disciplines or tasks, e.g., third-person singular present verbs. This explains the performance drop of the classifiers on Task 3.
\begin{figure}[h]
\centering
\vspace{-1mm}
    \includegraphics[width=0.85\columnwidth]{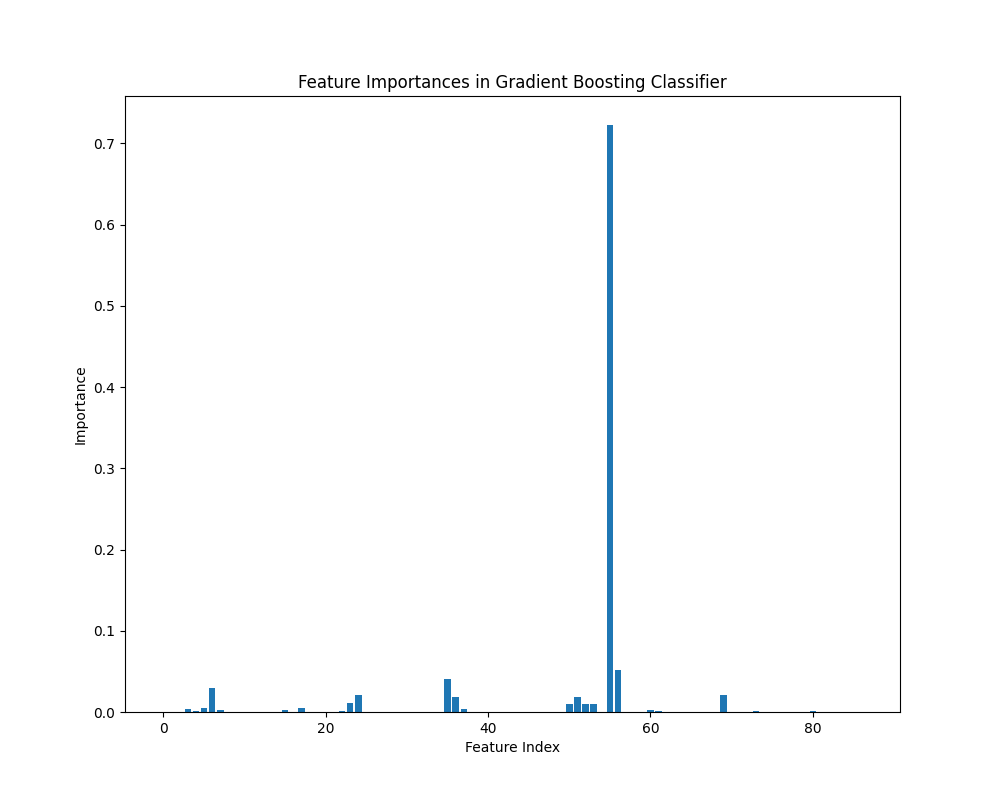}\\ 
    (a) CS, Task1. \\
    \includegraphics[width=0.85\columnwidth]{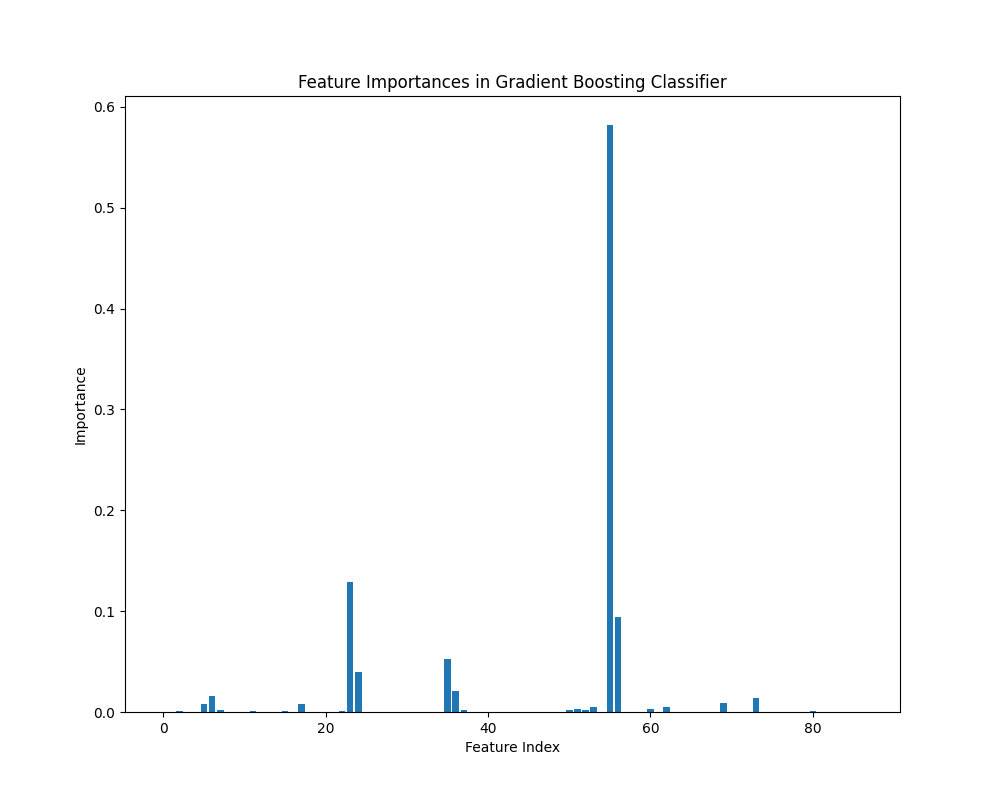} \\    
    (b) PHX, Task1. \\
    \includegraphics[width=0.85\columnwidth]{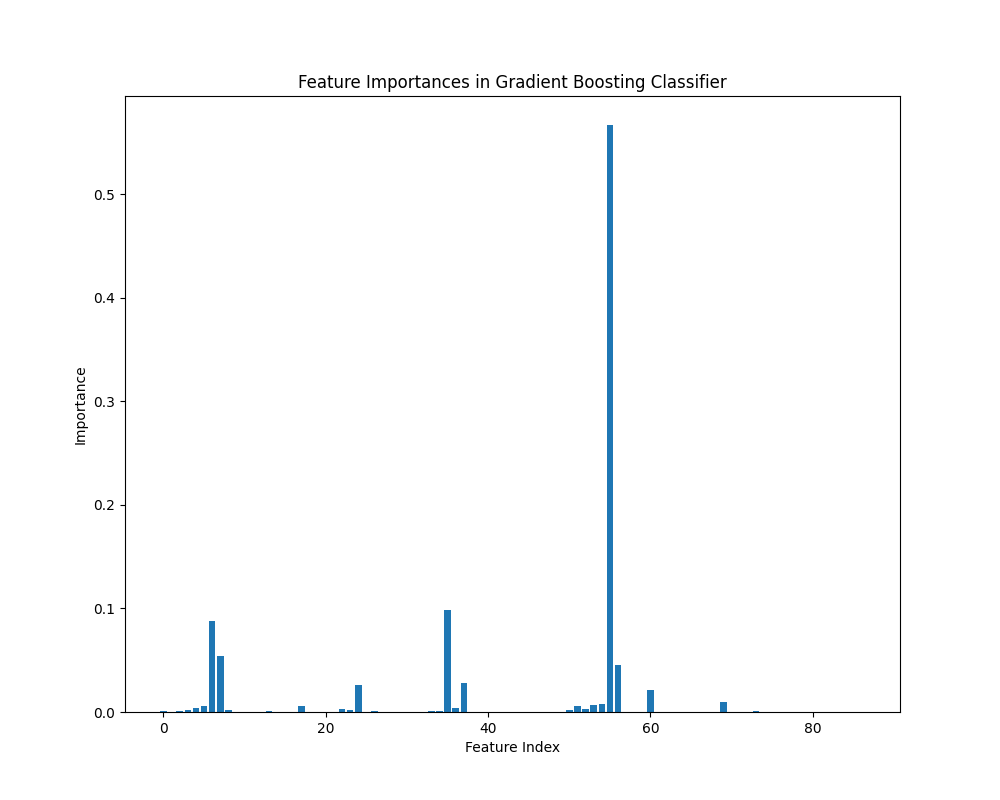} \\ 
    (c) HSS, Task1. \\
\caption{NELA feature importance for Task 1. Readability is the dominant feature for all three tasks.}
\label{fig:nelatask1}\vspace{-5mm}
\end{figure}

\begin{figure}[h]
\centering
\vspace{-1mm}
    \includegraphics[width=0.85\columnwidth]{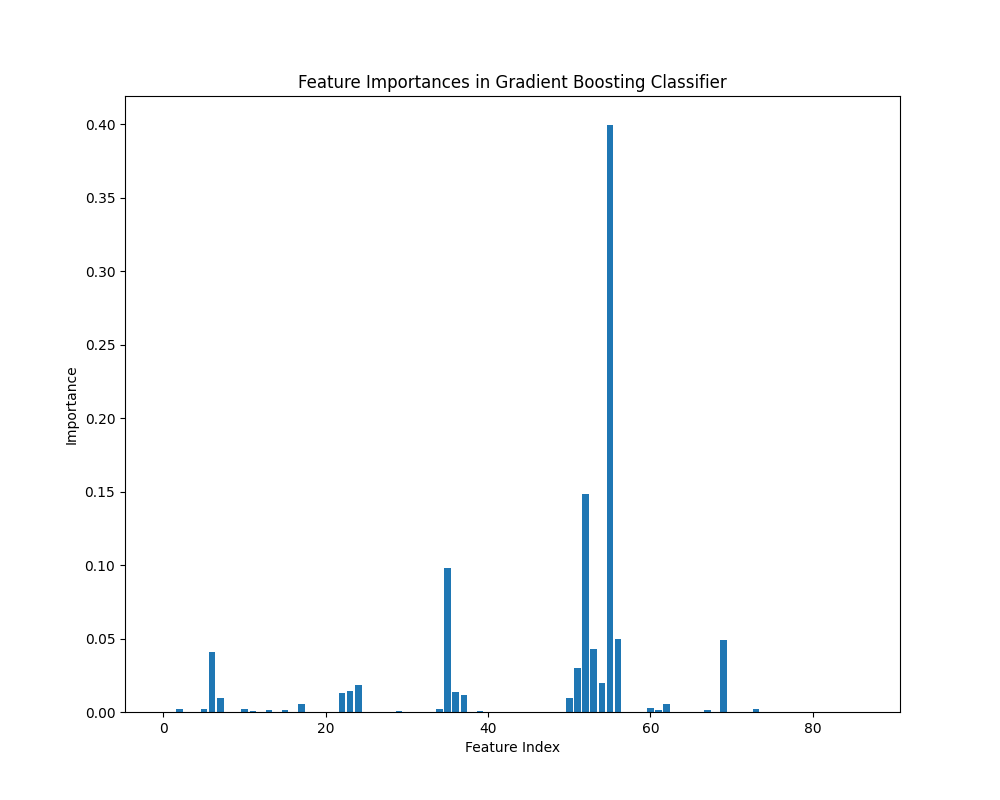}\\ 
    (a) CS, Task2. \\
    \includegraphics[width=0.85\columnwidth]{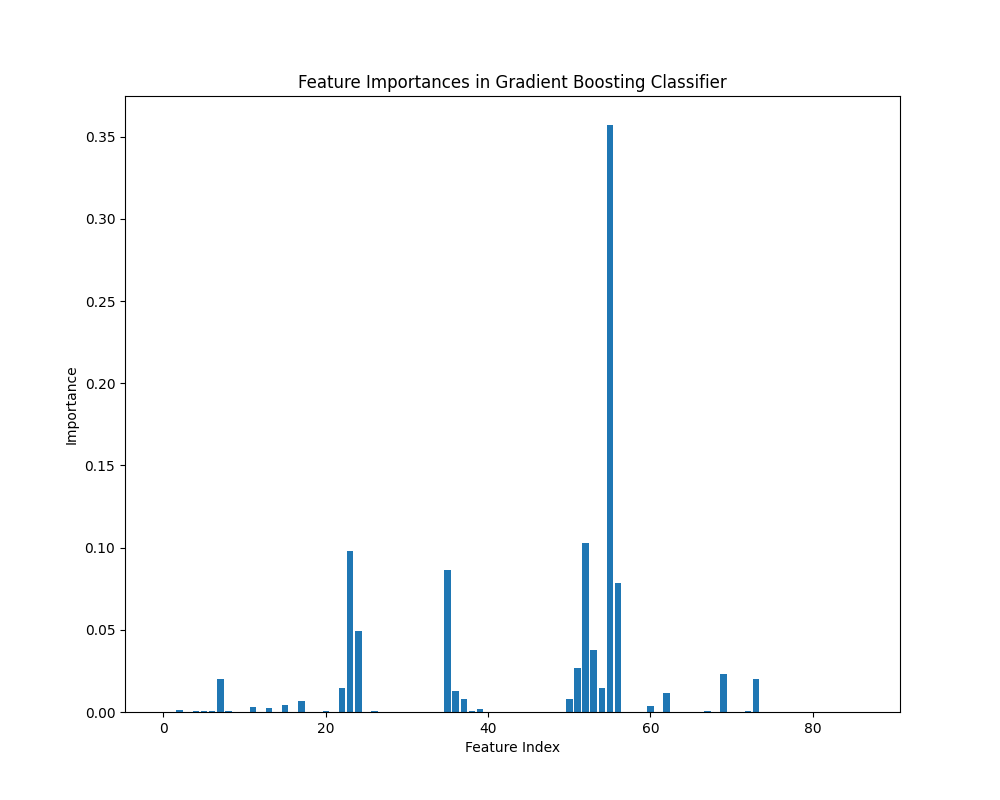} \\    
    (b) PHX, Task2. \\
    \includegraphics[width=0.85\columnwidth]{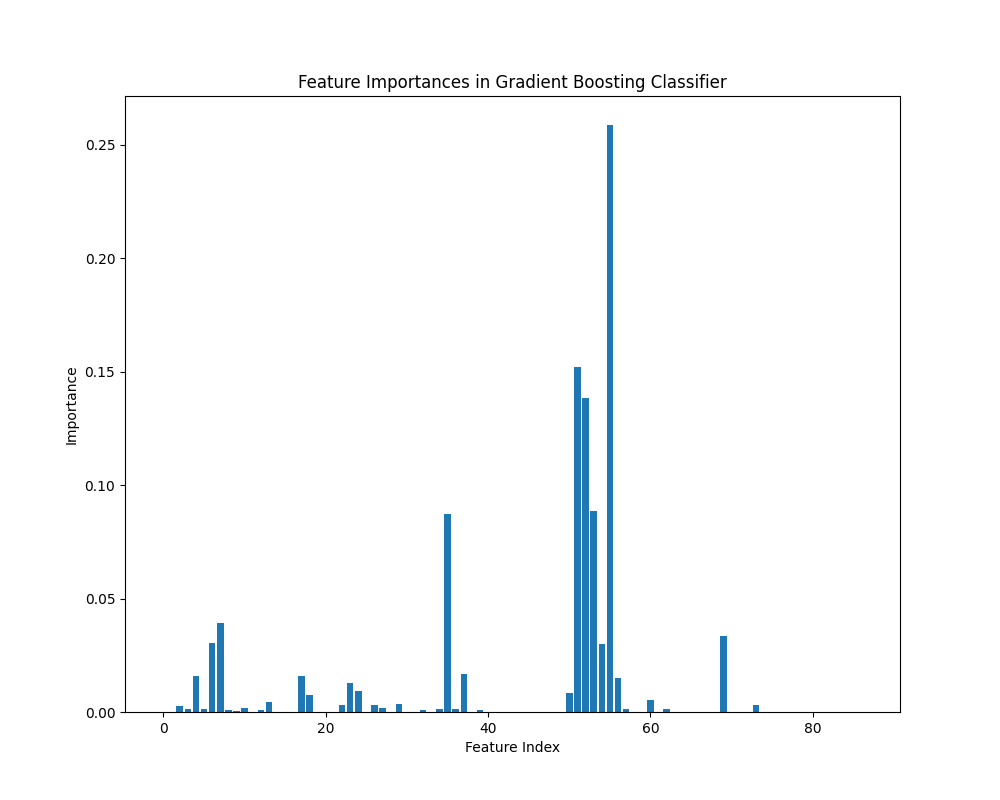} \\ 
    (c) HSS, Task2. \\
\caption{NELA feature importance for Task 2. More features get involved in the decision.}
\label{fig:nelatask2}\vspace{-5mm}
\end{figure}

\begin{figure}[h]
\centering
\vspace{-1mm}
    \includegraphics[width=0.85\columnwidth]{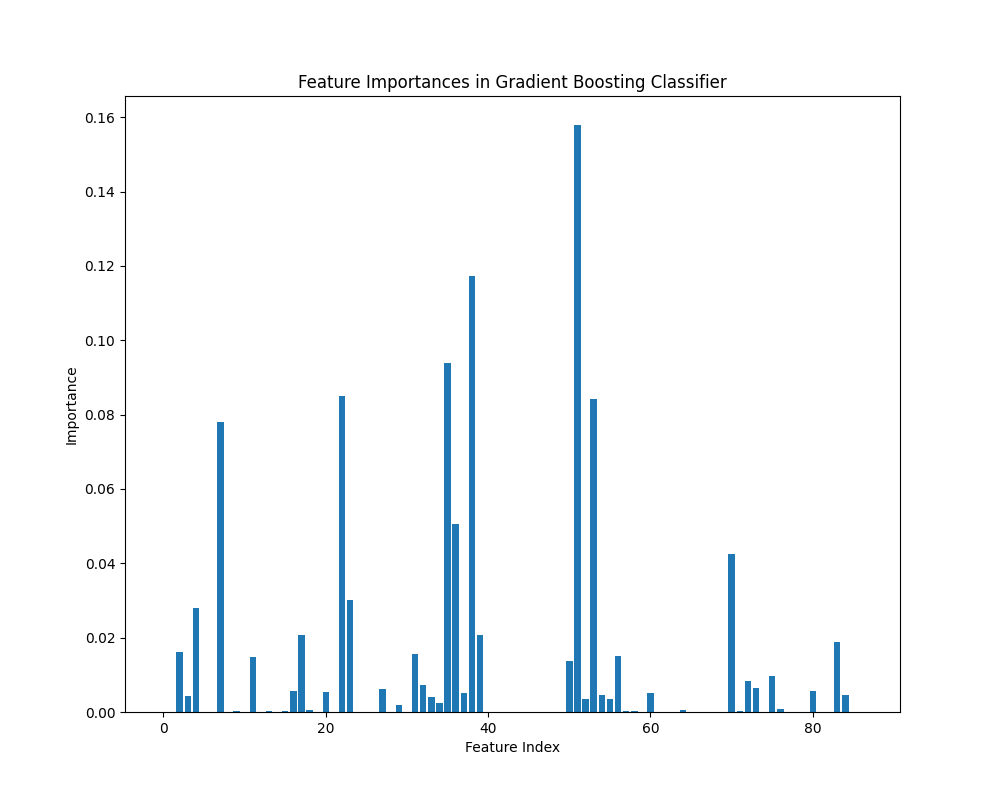}\\ 
    (a) CS, Task3. \\
    \includegraphics[width=0.85\columnwidth]{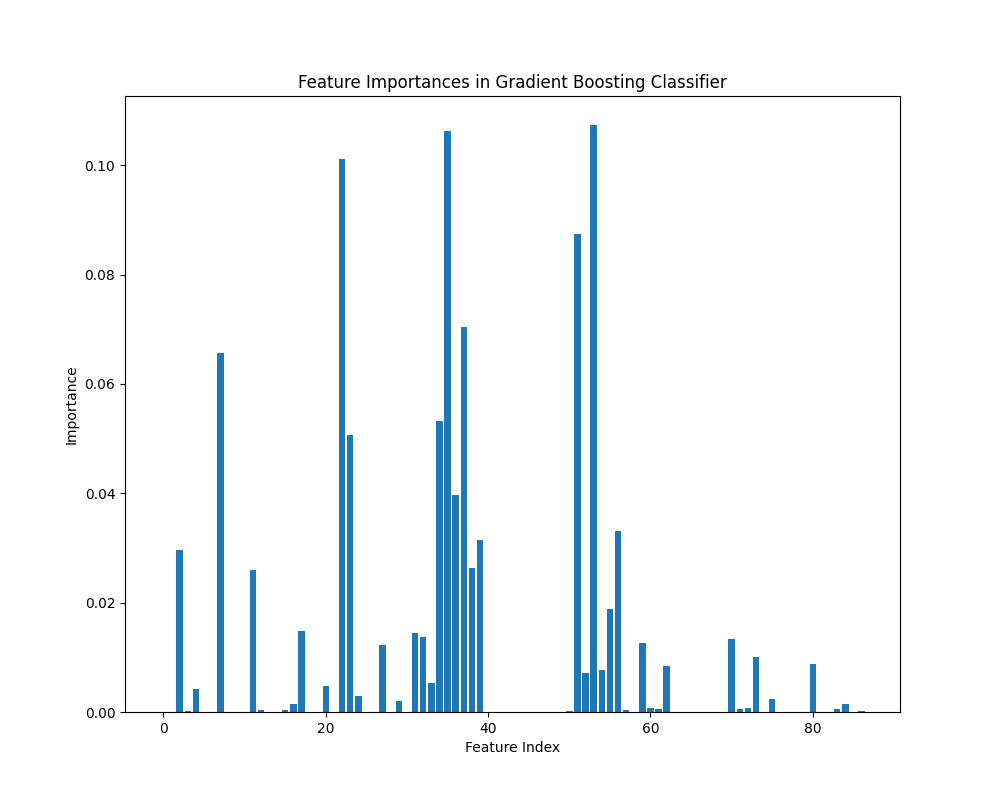} \\    
    (b) PHX, Task3. \\
    \includegraphics[width=0.85\columnwidth]{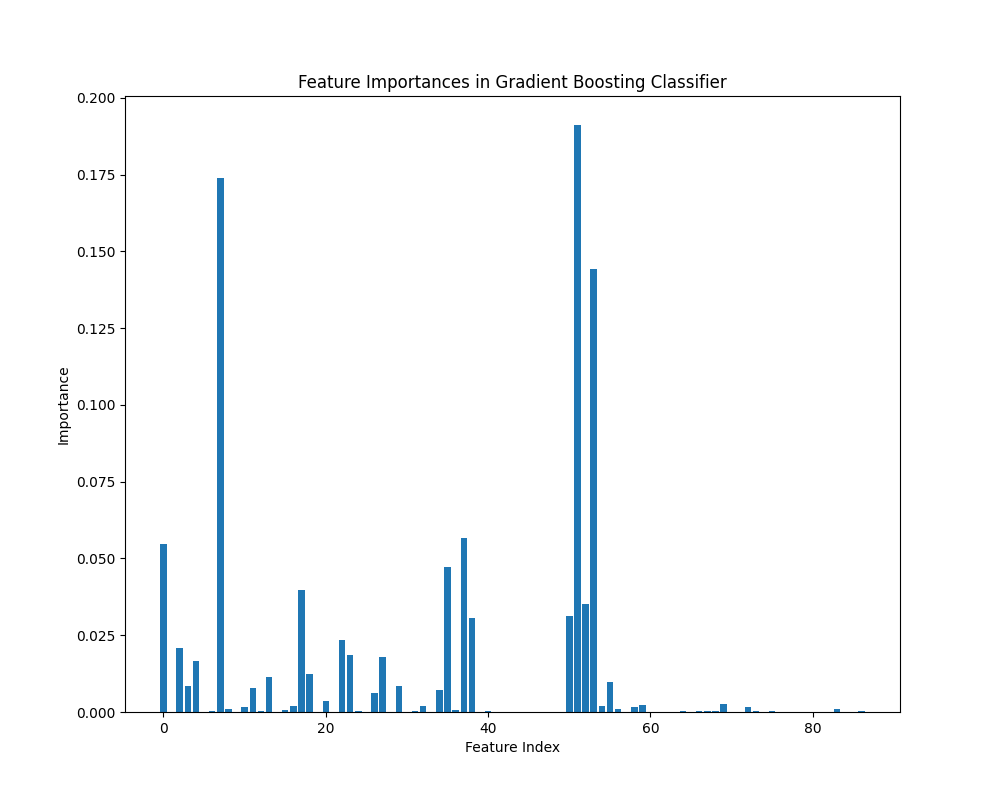} \\ 
    (c) HSS, Task3. \\
\caption{NELA feature importance for Task 3. Multiple features contribute to the decision.}
\label{fig:nelatask3}\vspace{-5mm}
\end{figure}

}

\subsection{Advanced Prompt Engineering} \label{apdx:promptsingle}
In Table~\ref{tab:morepromptsingle}, we show the testing accuracy of prompt-specific models on the advanced prompts. As discussed in Section \ref{sec:discussion}, the prompt-specific models perform worse than the cross-prompt models. Our interpretation is that the prompt-specific models may have learned some prompt-specific bias, i.e., linguistic features that are only generated by certain prompts. Meanwhile, the cross-prompt models are more likely to learn ChatGPT-specific features, i.e., features that consistently appear in ChatGPT-generated content from different prompts. 

\subsection{Rephrasing and Prompt Engineering Attacks} \label{apdx:propmtattack}
For rephrasing attacks, we provide the initial GPT-generated abstracts to ChatGPT. We ask it to rephrase it for two extra rounds. Our prompt is "1. Here is a paper abstract generated by ChatGPT. Please rewrite it to make it look more like a human-written abstract. {abstract}. 2. Please again rewrite the rewritten abstract given in your previous answer. Please only include the abstract in your response."

For prompt engineering attacks, we ask ChatGPT to perform Task 1, 2, and 3 but avoid using the Top-10 frequent vocabulary. We adhere "Also, please avoid using the following words and phrases: {vocabulary}" to the original prompts used in GPABench2.

\subsection{Copyediting Attacks} \label{apdx:ceaattack}
We present the unique vocabulary as follows. We use blue color to annotate the unique vocabulary of human writers used for substitution. For the replacement, we also consider different cases and tenses.

The Top-10 1-gram unique vocabulary for human-written abstracts is [{\newc 'article'}, 'with', 'here', 'may', 'there', 'find', {\newc 'consider'}, 's', 'using', 'possible', 'from', {\newc 'present'}, 'i', 'case', {\newc 'discuss'}, 'be', 'about', 'then', {\newc 'people'}, 'very', 'more', 'are', 'first', {\newc 'many'}, 'was', 'were', 'it', 'than', {\newc 'show'}, 'so', 'use', 'an', 'been', 'used', 'is', 'all', 'work', 'can', 'some', 'problem', 'e', {\newc 'we'}, 'as', {\newc 'particular'}, 'that', 'if', 'have', 'at', 'has', 'one', 'also', 'they', 'no', 'shown', 'most', 'not', 'or', 'what', 'but', 'finally', 'other', 'which', 'two']. The Top-10 bi-gram vocabulary for human-written abstracts is ['here we', 'we use', 'of a', 'number of', 'from the', {\newc 'in particular'}, 'and to', 'by the', 'declaration of', 'do not', 'based on', 'approval statement', 'with the', 'may be', 'has been', 'it is', {\newc 'we also'}, 'ethics approval', 'and the', 'than the', {\newc 'we discuss'}, 'and a', 'we prove', {\newc 'this work'}, 'in which', 'i e', 'show that', 'that the', 'can be', 'is shown', 'for the', 'be used', 'consistent with', 'results show', 'one of', {\newc 'we consider'}, {\newc 'study the'}, {\newc 'a new'}, 'there is', 'and that', 'in the', 'to be', 'in this', 'of the', 'of interests', 'is an', 'find that', 'as well', {\newc 'we present'}, 'is a', 'if the', 'we find', 'e g', 'is also', {\newc 'we show'}, 'finally we', 'that is', {\newc 'this article'}, {\newc 'we study'}, 'in a', 'to the', 'used to', 'effect of', 'paper we', 'with a', 'is the', 'the same', 'which is', 'at the'].

The Top-10 1-gram unique vocabulary for GPT-GEN abstracts (combining different prompts, tasks, and discilines) is {'insights', 'aims', 'explores', 'investigate', 'research', 'demonstrate', 'humanities', 'drawing', 'significant', 'contributions', 'computer', 'on', 'by', 'introduces', 'this', 'specifically', 'techniques', 'field', 'approach', 'into', 'additionally', 'science', 'using', 'authors', 'employing', 'light', 'comprehensive', 'innovative', 'novel', 'these', 'address', 'i', 'various', 'such', 'theoretical', 'paper', 'study', 'effectiveness', 'our', 'furthermore', 'individuals', 'presents', 'impact', 'potential', 'social', 'contributes', 'valuable', 'properties', 'sciences', 'examines', 'understanding', 'conducted', 'analysis', 'findings', 'through', 'focuses', 'proposed'}. The Top-10 bi-gram unique vocabulary for GPT-GEN abstracts is {'a comprehensive', 'the potential', 'paper introduces', 'aims to', 'the authors', 'valuable insights', 'the effectiveness', 'propose a', 'in this', 'paper we', 'of these', 'within the', 'the researchers', 'social sciences', 'analysis of', 'we demonstrate', 'to address', 'we propose', 'on the', 'additionally we', 'our understanding', 'our findings', 'paper presents', 'our analysis', 'impact of', 'paper examines', 'our approach', 'contributes to', 'examines the', 'of our', 'the impact', 'computer science', 'paper focuses', 'the findings', 'to the', 'the paper', 'the study', 'of this', 'and social', 'this paper', 'the field', 'contribute to', 'insights into', 'investigate the', 'this study', 'understanding of', 'this research', 'the innovative', 'additionally the', 'our research', 'the proposed', 'explores the', 'demonstrate that', 'light on', 'demonstrate the', 'in various', 'effectiveness of', 'address this', 'study we', 'shed light', 'novel approach', 'such as', 'focuses on', 'a novel', 'into the', 'presents a', 'paper i', 'field of', 'humanities and'}.

The Top-5 1-gram unique vocabulary for GPT-GEN abstracts is {'light', 'techniques', 'aims', 'humanities', 'our', 'explores', 'demonstrate', 'insights', 'paper', 'these', 'effectiveness', 'this', 'furthermore', 'significant', 'understanding', 'address', 'findings', 'sciences', 'additionally', 'study', 'comprehensive', 'through', 'focuses', 'potential', 'by', 'into', 'research', 'examines', 'various', 'approach', 'presents', 'novel', 'computer', 'innovative'}. The Top-5 bi-gram unique vocabulary for GPT-GEN abstracts is {'paper presents', 'the study', 'paper we', 'a novel', 'focuses on', 'this paper', 'insights into', 'computer science', 'understanding of', 'light on', 'to address', 'the findings', 'a comprehensive', 'the proposed', 'humanities and', 'impact of', 'the effectiveness', 'such as', 'this study', 'explores the', 'our findings', 'on the', 'the field', 'and social', 'into the', 'the paper', 'effectiveness of', 'in this', 'our research', 'contribute to', 'examines the', 'paper i', 'social sciences', 'this research', 'aims to', 'valuable insights', 'demonstrate the', 'additionally we', 'demonstrate that', 'of our', 'paper focuses', 'the authors'}.

The Top-3 1-gram unique vocabulary for GPT-GEN abstracts is {'effectiveness', 'aims', 'examines', 'by', 'novel', 'into', 'various', 'our', 'humanities', 'research', 'light', 'through', 'findings', 'approach', 'innovative', 'understanding', 'comprehensive', 'these', 'additionally', 'this', 'potential', 'insights', 'paper', 'furthermore', 'demonstrate', 'focuses'}. The Top-3 bi-gram unique vocabulary for GPT-GEN abstracts is {'contribute to', 'our findings', 'the field', 'to address', 'this study', 'on the', 'a novel', 'insights into', 'such as', 'demonstrate that', 'examines the', 'aims to', 'paper presents', 'a comprehensive', 'the study', 'understanding of', 'this research', 'light on', 'our research', 'this paper', 'the authors', 'social sciences', 'paper i', 'the paper', 'in this', 'into the', 'the effectiveness', 'focuses on', 'effectiveness of', 'demonstrate the'}.

The Top-10 set of replacement pairs is [('insight', 'grasp'), ('contribution', 'improvement'), ('novel', ''), ('theoretical', ''), ('effectiveness', 'accuracy'), ('specifically', 'in particular'), ('additionally', 'also'), ('innovative', ''), ('valuable', ''), ('address', 'solve'), ('impact', 'influence'), ('significant', ''), ('conduct', 'do'), ('approach', 'method'), ('furthermore', 'besides'), ('analysis', 'study'), ('understanding', 'grasp'), ('various', 'many'), ('comprehensive', ''), ('individuals', 'people'), ('field', 'subject'), ('property', 'feature'), ('employing', 'applying'), ('technique', 'method'), ('I', 'we'), ('drawing', 'getting'), ('finding', 'result'), ("a novel", "a new"), ('furthermore we', 'we also'), ('proposed', ''), ('explore', 'discuss'), ('examine', 'consider'), ('investigate', 'study'), ('introduce', 'present'), ('demonstrate', 'show '), ('shed light on', 'help explain'), ('such as', 'like'), ('the authors/researchers', 'we'), ('insights into ', 'grasps of '), ('understanding of ', 'grasps of '), ('our analysis of ', 'a study of '), ('the field of computer science/physics/social sciences/humanities and social sciences', 'this subject'), ('the/a/its potential', 'the/a/its possibility'), ('our/this paper/research/study/work/article aims to', 'our goal is to'), ('we aim to ', 'our goal is to '), ('focus on ', 'discuss '), ('our/this/the research/paper/study', 'we/this work/this article'), ("In this paper /work/article/study/research", "")].

The Top-5 set of replacement pairs is [('insight', 'grasp'), ('novel', ''), ('effectiveness', 'accuracy'), ('additionally', 'also'), ('innovative', ''), ('valuable', ''), ('address', 'solve'), ('impact', 'influence'), ('significant', ''), ('approach', 'method'), ('furthermore', 'besides'), ('understanding', 'grasp'), ('various', 'many'), ('comprehensive', ''), ('field', 'subject'), ('technique', 'method'),('finding', 'result'), ("a novel", "a new"), ('furthermore we', 'we also'), ('proposed', ''), ('explore', 'discuss'), ('examine', 'consider'), ('demonstrate', 'show '), ('shed light on', 'help explain'), ('such as', 'like'), ('the authors', 'we'), ('insights into ', 'grasps of '), ('understanding of ', 'grasps of '), ('the field of computer science/social sciences/humanities and social sciences', 'this subject'), ('the/a/its potential', 'the/a/its possibility'), ('our/this paper/research/study/work/article aims to', 'our goal is to'), ('we aim to ', 'our goal is to '), ('focus on ', 'discuss '), ('our/this/the research/paper/study', 'we/this work/this article'), ("In this paper/work/article/study/research", "")].

The Top-3 set of replacement pairs is [('insight', 'grasp'), ('novel', ''), ('effectiveness', 'accuracy'), ('additionally', 'also'), ('innovative', ''), ('approach', 'method'), ('furthermore', 'besides'), ('understanding', 'grasp'), ('various', 'many'), ('comprehensive', ''), ('field', 'subject'), ('finding', 'result'), ("a novel", "a new"), ('proposed', ''), ('examine', 'consider'), ('demonstrate', 'show '), ('shed light on', 'help explain'), ('such as', 'like'), ('the authors', 'we'), ('insights into ', 'grasps of '), ('understanding of ', 'grasps of '), ('the field of computer science/social sciences/humanities and social sciences', 'this subject'), ('the/a/its potential', 'the/a/its possibility'), ('our/this paper/research/study/work/article aims to', 'our goal is to'), ('we aim to ', 'our goal is to '), ('focus on ', 'discuss '), ('our/this/the research/paper/study', 'we/this work/this article')), ("In this paper/work/article/study/research", "")].



\section{Model Interpretation} \label{apdx:interpret}

\subsection{Model Interpretation}
Besides the accuracy, the transparency of \checkgpt~is also important. 
The interpretation of \checkgpt~not only helps us understand the rationale behind a specific decision but also provides discerning insights to distinguish AI-generated from human-written texts. 
Therefore, to investigate this, we employ two methods: Integrated Gradients~\cite{sundararajan2017axiomatic, kokhlikyan2020captum} and Shapley Values~\cite{datta2016algorithmic, lundberg2017unified}. They represent two different angles: model-specific and model-agnostic explainability
\begin{itemize}[noitemsep, topsep=0mm, left=0pt]
\item \textbf{Integrated Gradients.} 
This method assigns the importance to each value by the gradients compared with the baselines along the path. The baselines are the inputs that induce a ``neutral'' decision. We utilize the implementation in~\cite{kokhlikyan2020captum} and apply it to our models.

\item \textbf{Shapley Values.} 
Originally introduced in ~\cite{shapley1953value} and recently applied to machine learning interpretation, 
a Shapley Value quantifies the impact of each feature by perturbing the input value and seeing how the change of input contributes to prediction. 
We adopt the implementation in~\cite{lundberg2017unified}.
\end{itemize}

\begin{figure*}[b]
\centering
\vspace{-1mm}
\begin{tabular}{c}
    \includegraphics[width=1.0\textwidth]{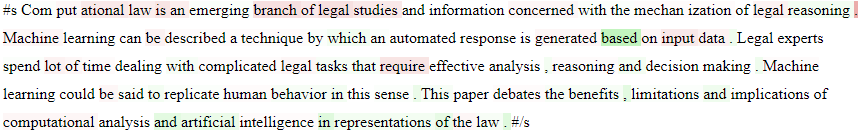}\\ 
    \vspace{3mm}(a) Human-written text. \\
    \includegraphics[width=1.0\textwidth]{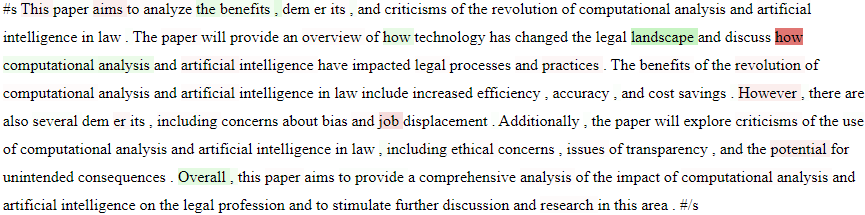} \\    
    \vspace{3mm}(b) GPT-written text in Task 1. \\
    \includegraphics[width=1.0\textwidth]{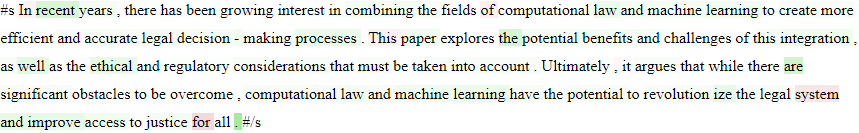} \\ 
    \vspace{3mm}(c) GPT-completed text in Task 2. \\
    \includegraphics[width=1.0\textwidth]{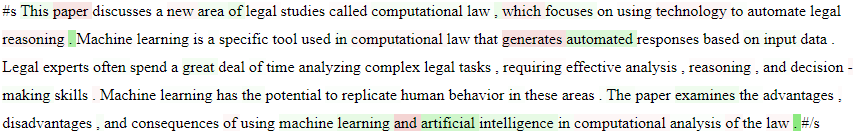} \\
    (d) GPT-polished text in Task 3. \\
\end{tabular}\vspace{-1mm}
\caption{Word importance using Integrated Gradients. A case of HSS written by humans and ChatGPT in three different tasks. Green regions indicate positive contributions to the corresponding label, and red ones indicate negative contributions.}
\label{fig:ig}\vspace{-5mm}
\end{figure*}

\vspace{2mm}
\noindent\textbf{Word-level Analysis.} 
We first apply these methods at the word level to measure the contribution of each word toward the decisions.
As the example shown in Fig.~\ref{fig:ig} using Integrated Gradients, the word ``landscape'' and ``automated'' are identified as the most significant features for Task 1 and Task 3, respectively. The feature saliency is almost uniformly distributed across the entire paragraph in Task 2. 

\begin{figure*}[t]
\centering
\vspace{-1mm}
\begin{tabular}{c}
    \includegraphics[width=1.0\textwidth]{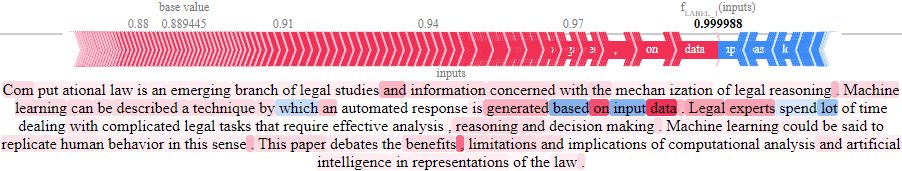}\\
    \vspace{3mm}(a) Human-written text. \\
    \includegraphics[width=1.0\textwidth]{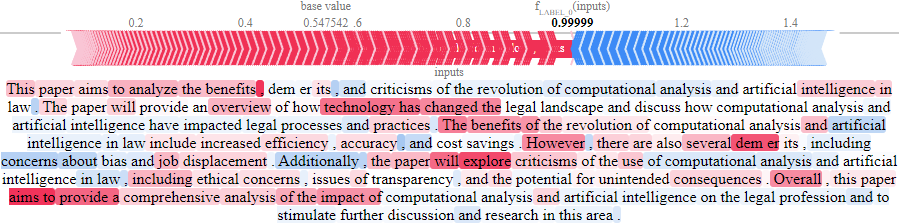} \\    
    \vspace{3mm}(b) GPT-written text in Task 1. \\
    \includegraphics[width=1.0\textwidth]{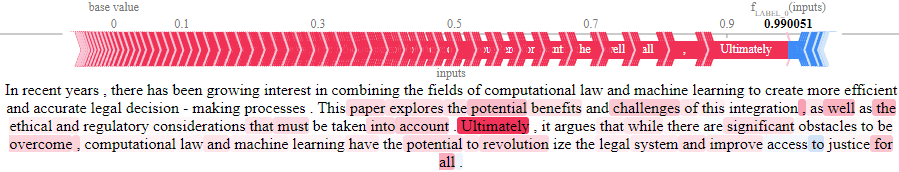} \\ 
    \vspace{3mm}(c) GPT-completed text in Task 2. \\
    \includegraphics[width=1.0\textwidth]{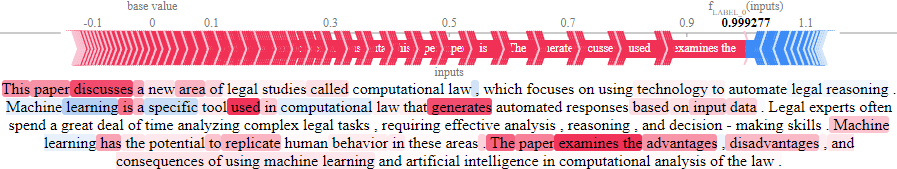} \\ 
    (d) GPT-polished text in Task 3. \\
\end{tabular}\vspace{-1mm}
\caption{Word importance using Shapley Values. A case of HSS written by humans and ChatGPT in three different tasks. Red regions indicate positive contributions to the label of a particular text, while blue ones indicate negative contributions.}
\label{fig:shap}\vspace{-5mm}
\end{figure*}

Fig~\ref{fig:shap} shows the comparisons of GPT-generated abstract and human-written abstract explained by Shapley Values. The words ``on'' and ``data'' are the most supportive features leading to a decision of ``human-written''. Words of metadiscourses are the most important features in Task 1 and Task 3. Reporting verbs like ``explore'', ``aim'', ``discuss'' and ``examine'' are mostly adopted by the ``GPT-written'' style for describing intentions. The transitional phases that guide the readers, like ``However'', ``Overall'' and ``Ultimately'', are also significant features for GPT writing, especially Task 2.

Our attempts at the word-level experiments produce relatively \textit{uninformative} findings. The significance assigned to each individual word is usually insufficient for human users to draw useful conclusions. The limitation of the methods is due to the sophistication of the LLMs which capture complex semantic and linguistic features. Thus, the word-level interpretations are inadequate for our analysis.

\vspace{2mm}
\noindent\textbf{Sentence-level Analysis.} While independent words do not show a sufficient power of explainability, the corporative semantic patterns captured within sentences have the potential to give a more comprehensive insight.
Language comprehension relies heavily on context, nuance, and syntactic structures, which are far more informative beyond interpreting individual words. 
Furthermore, the LLMs like ChatGPT, typically build their task to generate coherent and sentence-level responses. Thus, a sentence-level analysis has been conducted in the hope of a better-quality interpretation. 

\begin{figure*}[t]
\centering
\vspace{-1mm}
\begin{tabular}{c}
    \includegraphics[width=1.0\textwidth]{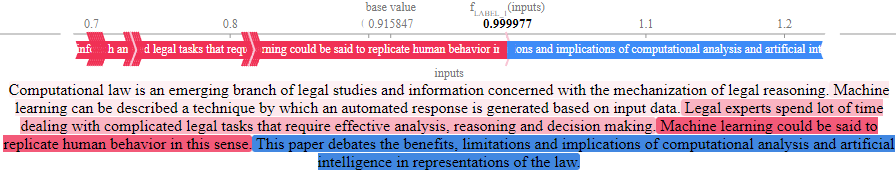}\\ 
    \vspace{3mm}(a) Human-written text. \\
    \includegraphics[width=1.0\textwidth]{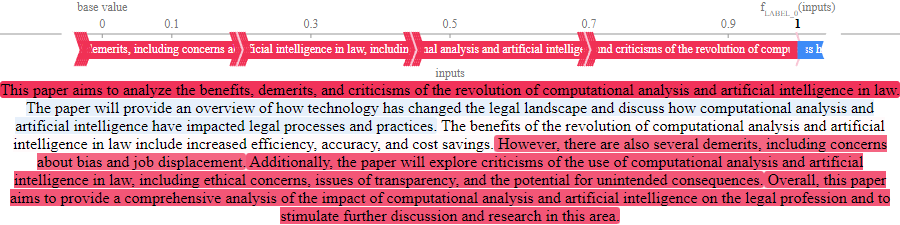} \\    
    \vspace{3mm}(b) GPT-written text in Task 1. \\
    \includegraphics[width=1.0\textwidth]{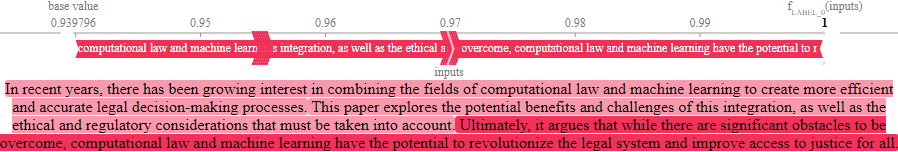} \\ 
    \vspace{3mm}(c) GPT-completed text in Task 2. \\
    \includegraphics[width=1.0\textwidth]{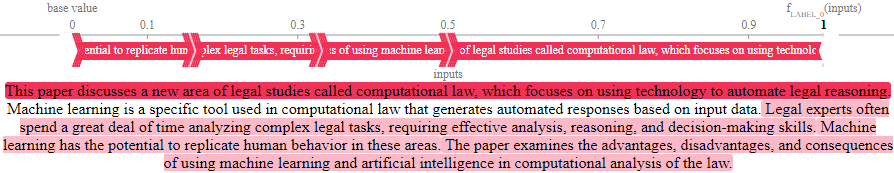} \\
    (d) GPT-polished text in Task 3. \\
\end{tabular}\vspace{-1mm}
\caption{Sentence importance using Shapley Values. A case of HSS written by humans and GPT in three different tasks. Red regions indicate positive contributions to the label of a particular text, while blue ones indicate negative contributions.}
\label{fig:shapsen}\vspace{-5mm}
\end{figure*}

In Fig~\ref{fig:shapsen}, we extend our analysis to sentence-level interpretations using Shapley Values because of its coherent output.
The abstracts are parsed into sentences, which become the new units of features.
The results show that sentence-level analysis provides more meaningful and consistent insights for identifying GPT-generated texts. 
First, we find that the supporting sentences for human texts or GPT texts are located differently in an abstract, which means that the GPT writing style for different presentation goals contains distinguishable and unique patterns for detection.
Second, we can see that ChatGPT frequently starts the abstract with a declarative statement like ``This paper proposes" to emphasize the focus of the paper. It shows that the particular ways of presenting ideas consist of another character of GPT's ``footprint'' in writing. 
Last, in the last sentence of the abstracts, ChatGPT usually tries to use a conclusive statement to summarize the findings or contributions of the paper, which is also widely observed in regular ChatGPT conversations. This ``habit'' of summarization, which is designed for Q\&A tasks, reveals the third pattern uniquely carried by GPT even when it is writing abstracts.
Additional examples of our interpretations are given in the Appendix.

In summary, comparing the interpretations derived from word- and sentence-level results, we find that the complex linguistic and presentation patterns can be better expressed by sentence-level features. However, we must admit that it also trades off the granularity and thus currently can not provide results in finer details (e.g., patterns of wording and phrasing).
These interpretation experiments demonstrate that no explicit or dominant indicators can be easily captured for GPT writing recognition. The finding emphasizes the necessity for applying sophisticated and automated tools, like deep learning techniques, to perform effective detection for complicated and subtle semantic features. 


